\pdfoutput=1
\def\year{2021}\relax
\documentclass[letterpaper]{article} 
\usepackage{aaai21}  
\usepackage{times}  
\usepackage{helvet} 
\usepackage{courier}  
\usepackage[hyphens]{url}  
\usepackage{graphicx} 
\usepackage{natbib}  
\usepackage{caption} 
\usepackage{amsmath}
\usepackage{subfig}
\usepackage{verbatim}

\usepackage{cite}
\usepackage{soul}
\usepackage[switch]{lineno}
\usepackage{xcolor}
\usepackage{amsbsy}

\usepackage[normalem]{ulem}
\newcommand{\stkout}[1]{\ifmmode\text{\sout{\ensuremath{#1}}}\else\sout{#1}\fi}

\title{User Driven Model Adjustment via Boolean Rule Explanations}

\title{User Driven Model Adjustment via Boolean Rule Explanations}
\author {
    Elizabeth M. Daly, Massimiliano Mattetti, \"{O}znur Alkan, Rahul Nair \\
}
\affiliations{
    IBM Research\\
    Dublin, Ireland
 
}

\usepackage[linesnumbered,ruled,vlined]{algorithm2e}

\SetKwInput{KwInput}{Input}
\SetKwInput{KwOutput}{Output}

\begin{document}


\maketitle

\begin{abstract}
AI solutions are heavily dependant on the quality and accuracy of the input training data, however the training data may not always fully reflect the most up-to-date policy landscape or may be missing business logic. The advances in explainability have opened the possibility of allowing users to interact with interpretable explanations of ML predictions in order to inject modifications or constraints that more accurately reflect current realities of the system. In this paper, we present a solution which leverages the predictive power of ML models while allowing the user to specify modifications to decision boundaries. Our interactive overlay approach achieves this goal without requiring model retraining, making it appropriate for systems that need to apply instant changes to their decision making. We demonstrate that user feedback rules can be layered with the ML predictions to provide immediate changes which in turn supports learning with less data. 
\end{abstract}

\section{Introduction}
AI is increasingly integral in many real world tasks from loan approval to forecasting organizational revenue. While supervised ML tasks such as classification perform well with appropriate historical data, they can not be updated quickly to support more dynamic situations. Retraining can be costly, and more challenging still is obtaining labeled data that accurately reflects the current decision landscape. Consider a loans approval application where there exists a business policy that a loan request should be accepted for any user with \(age>30 \land income>50k \land education=``Masters"\) and a machine learning model has been trained based on historical data to predict if a user will be approved or not by leveraging many other features other than the three mentioned in the business logic. Now consider a new policy comes into place where the rule changes from age$>$30 to age$>$26. Based on historical data the machine learning model may incorrectly reject users between the ages 26-30. 
Options include relabelling historical data in order to accurately reflect the new decision boundary which is labour intensive. Alternatively, the solution must wait for new data to be collected with the correct labels. In both cases, the model must be retrained after the data has been updated. These approaches can result in a lack of data for some instances' coverage and may be time consuming making it less appropriate for applications where decision processes or policies may change.



\begin{figure*}[t]
\captionsetup[subfigure]{justification=centering}
\centering
\subfloat[Problem]{
\includegraphics[width=.5\columnwidth]{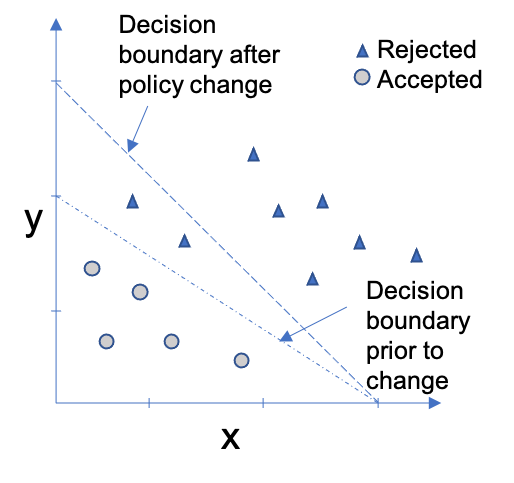}
}
\subfloat[ML Approach:Relabel and retrain][ML Approach:\\Relabel and train]{
\includegraphics[width=.5\columnwidth]{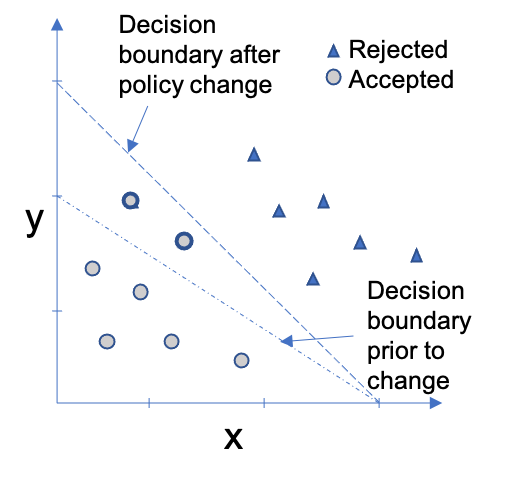}
}
\subfloat[ML Approach:Collect new data and train][ML Approach:\\Collect new data and train]{
\includegraphics[width=.5\columnwidth]{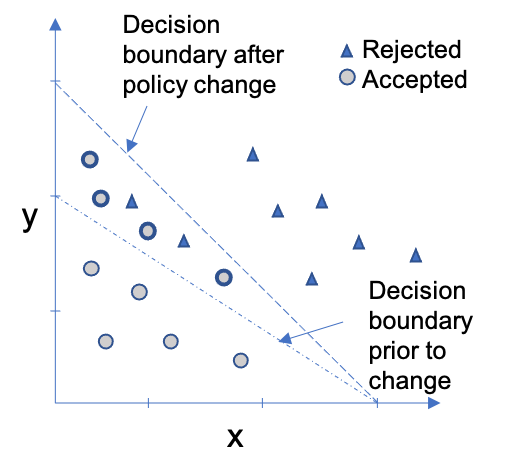}
}
\subfloat[Interactive Overlay\\Approach]{
\includegraphics[width=.5\columnwidth]{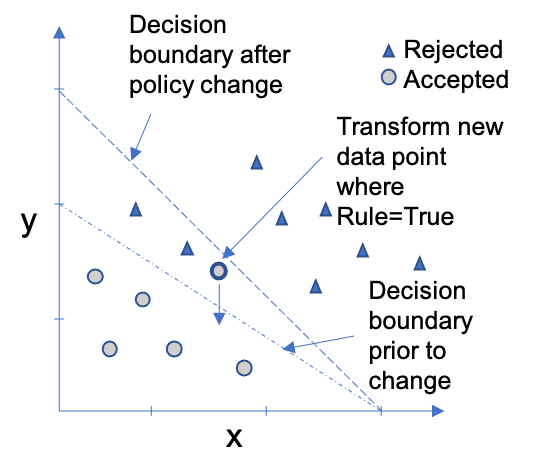}
}

\caption{Approaches to learning new decision boundaries}
\label{fig:mllearning}
\end{figure*}

In this paper, we investigate the scenario where the data used to train an existing ML model may not reflect the current decision criteria. Additionally, the data may be lacking external knowledge such as policies or policy changes. In order to tackle this challenge, we present a solution that leverages the predictive power of an ML model with user contributed 
decision rules to enable adjustments online acting as a ``patch" or ``bandaid" taking immediate affect. Figure \ref{fig:mllearning} illustrates how current ML solutions cope with changing decision boundaries in a simplified linear scenario: the solution is either to relabel existing data, or wait until sufficient new data is available. Active learning has shown promise in reducing the amount of newly labelled data needed, however model retraining is still required and in practise users may need to label many data instances in order to impact the model, which can prove frustrating for the user \cite{cakmak2010designing, guillory2011simultaneous}. In an online scenario where policies need to have an immediate effect, for example in a spam detection setting, waiting for relabelling and retraining to update the model may not be possible. Our solution attempts to harness the underlying ML model by storing a series of user adjustments in the form of decision rules that can be applied to push future instances to the appropriate decision boundary in order to reflect the user constraints. We present this solution as a user modifiable layer that can support immediate changes and influence over an existing model. Our goal is to use an existing ML model, while permitting users to provide rule based modifications that adjust the final decision making criteria allowing them to adjust the predictions for specified parameters. This is achieved without modifying or updating the underlying ML model but by post-processing the model output through a combination of interpretable Boolean rules and inferred transformation functions which modify the input request to reflect the user feedback. 

One question might arise: why not either use a complete rule based system or simply create a post processing filtering layer. However, a user may not be able to specify and define the entire decision criteria, whereas they can easily contribute corrections or updates that are reflective of a number of key variables. Consider our example for a loan scenario where many different features are factors in the underlying prediction and the only change a user wants to make is to adjust the age requirement from 30 to 26. For a rule system to achieve the same level of accuracy as an ML model the rules may need to become increasingly complex which can hurt interpretability \cite{Interpretable_Decision_Sets_KDD2016}. In the case of a post processing filter, the solution would not necessarily know that the primary contributing factor for the label was indeed the age and therefore would need to do some post analysis to determine the appropriate label. 

To tackle these challenges, the rule-based solution we propose does not try to capture all rules and features but makes a trade off between compact interpretable rules that are as accurate as possible while the underlying decision making is still governed by the ML model. As a result, only minor adjustment on known features or thresholds are needed and understanding the entire feature space or increasingly complex rules is not needed. Our solution provides a mechanism to support post analysis in a generalizable way, by storing user post filtering rules and then adjusting the input instance to determine if the variable the user wishes to adjust is the contributing factor. Additionally, it is important to note that in many circumstances we do not expect the interactive overlay approach to outperform an ML model with access to sufficient correctly labeled data. 

In this paper, we aim to address the following research questions: R1) Can we create an explainable interactive overlay that supports modifications to an existing ML model decision process without retraining the model? R2) Can our interactive overlay system with only partial knowledge converge to the same performance as an ML model with full knowledge through rule based modifications? The rest of the paper is organized as follows. We first review related work and then present our solution for supporting modifications to an existing ML model through user feedback. We evaluate our framework with benchmark datasets and finally, we conclude the paper highlighting future lines of research.


\section{Related Work}



Explainability has become an increasing area of interest in the AI community given the number of high stake scenarios where machine learning is employed. The goal of explainable AI is to make the decision making process understandable to a user \cite{gunning2017explainable}. Many different approaches have been proposed, some of which involve new predictive algorithms where explainability is built in, and others focus on post-hoc explanations agnostic to the underlying algorithm. These techniques include calculating feature importance \cite{guidotti2018survey, ribeiro2016should}, finding similar instances from past predictions \cite{gurumoorthy2019efficient}, identifying what features are present or missing to support the prediction known as contrastive explanations \cite{dhurandhar2018explanations} or generating interpretable rule based representations \cite{dash2018boolean,ribeiro2018anchors}. Yousefzadeh and O'Leary present a mechanism for finding `flip' points which provides the smallest change to a single continuous feature that would induce a change to the output of the model  \cite{yousefzadeh2020auditing}. 

While improvements in explainability and interpretability may assist in allowing the user to trust an AI system, it does not allow the user to correct errors or add domain logic. In reality, many ML solutions deployed in applications require some level of business logic which is either crafted by the data scientist into the solution through data selection or as a set of post processing logic. The need for user control and interpretability has garnered a resurgence in rule based models \cite{Interpretable_Decision_Sets_KDD2016, dash2018boolean}. In \cite{vojivr2020editable} the authors present a solution which generates a rule-based system from data that the user may modify which is then used as a rule-based executable model. While rule based models have the advantage of being interpretable, in order to achieve coverage the model must add rules to cover increasingly narrow slices which can in turn negatively impact interpretability. Our work is complementary as the main focus is on a user editing rules whereas our goal is to build a framework to allow such modifications to adjust/patch a trained ML model where adjustments may require changes to a small subset of rules or clauses and the remaining decision process is still governed by the ML model.

Fails and Olson were the first to introduce the term Interactive Machine Learning (IML) \cite{fails2003interactive,amershi2014power}.  They presented an Interactive ML framework where the ML model is intentionally trained quickly and the results are presented to the user, allowing the user to give feedback, explore the impact of their changes and then tune their feedback accordingly. 


The most common way for a user to influence or instill knowledge in an AI system is to provide labels and active learning has been leveraged to reduce the load on users by intelligently selecting which data instances to present to the user for annotation. However, users may need to label many instances in order to impact the model, leading to labelling fatigue \cite{cakmak2010designing, guillory2011simultaneous}. 

The work most similar to our own combines active learning with explanations and user corrections. Teso et al. presented a framework for Explanatory Interactive Learning (XIL) \cite{teso2019explanatory}. Active learning is employed to select a data instance to present to the user along with the prediction and an explanation based on feature importance. If the prediction is incorrect the user is given the opportunity to relabel the data instance. If the prediction is correct but the explanation is incorrect the user can give feedback on the presented LIME values \cite{ribeiro2016should} and may flag a feature as irrelevant. This information is then used to generate counter example instances that are identical except the irrelevant feature is modified either through randomization or some other strategy. These new data instances are then used as input for future retraining to influence the model. Our work is complimentary to this space; however, we leverage the interpretable value of Boolean rules to allow a user to give much more fine-grained feedback. A similar mechanism of taking user feedback which modifies the input training data is presented in \cite{schramowski2020making}, however both solutions require model retraining for the user changes to be reflected in the model. Our solution enables modifications to take immediate effect, without model retraining, through post processing of data instances. 


\section{Preliminaries}

To simplify the discussions, we present our solution considering the binary classification problem, however, the ideas could be extended to multi-class classification problems as well. We consider a binary classification problem from a domain D(X,Y), from which $n$ i.i.d. samples are drawn ($x_{i},y_{i}$), $i \in \{1, ..., n\}$ with labels $y_{i} \in \{0, 1\}$. Below, we present the important definitions and terminology that we use to describe our solution.

\textbf{Rule}. Following the terminology from previous works \cite{dash2018boolean, Interpretable_Decision_Sets_KDD2016}, we define a \textit{rule} $R$ as a tuple $R(e,p)$, where $e$ is a boolean \textit{clause}, and $p$ is the class label assigned to all the instances satisfying $e$. A \textit{clause} is a conjunction of conditions over a subset of features in $X$.

\textbf{Condition}. A condition is defined as a triple \(\langle variable, comparison\:operator, value \rangle\) where the variable represents a feature in $X$ and the comparison operator can be one in the set \{`$=$',`$\neq$',`$>$',`$\geq$',`$<$',`$\leq$'\}. 

\textbf{Rule Satisfiability}. An instance $x_i$ satisfies a clause $e$ if all the conditions in $e$ are evaluated to $True$ on $x_i$. By extension, we say that an instance $x_i$ satisfies a rule $R(e,p)$ if $x_i$ satisfies the boolean clause $e$ in $R$. Formally, we define a boolean clause as a function $e: X \to \{0,1\}$, and an instance $x \in X$ \textbf{satisfies} $e$ $\iff$ $e(x) = 1$. 




\textbf{Feedback Rule}. A \textit{feedback rule} ($FR$) is defined as a triple of the form $FR(R,R',T)$, where $R$ contains the original rule, $R'$ contains the users feedback, which is the modified version of $R$ (with $e'$ and $p'$ indicating the boolean clause and the class label of the modified rule, respectively), and $T$ stores the transformation function.

\textbf{Conflicting Rules}. Two rules $R_1(e_1,p_1)$ and $R_2(e_2,p_2)$ are \textit{conflicting} if there exists at least one instance $x_i$ that satisfies $e_1$ and $e_2$ but $p_1 \neq p_2$. There are two types of conflicting rules to be considered. The first is where two rules from the underlying explainer cover the same data instance but link to a different class. For the purposes of this paper we will make the simplified assumption that the rules are \textit{conflict-free}. The second situation where conflicting rules may come into play is where a user modifies a rule which then results in a conflict of other feedback rules. In order to support this scenario a conflict analysis, such as the one presented in~\cite{lindgren2004methods}, would need to occur to allow the user to understand which other rules would be impacted and potentially need to be updated based on their feedback. For the purposes of this paper however we ensure feedback rules are \textit{conflict-free}.


\textbf{Transformation Function}. A transformation function is a function that modifies an input instance $x_i$ which previously satisfied a rule $R$ so as to turn it into an instance $x_{i}'$ that satisfies a modified rule $R'$.

\section{Proposed Interactive Overlay}

\begin{figure}
\centering
\includegraphics[width=1\columnwidth]{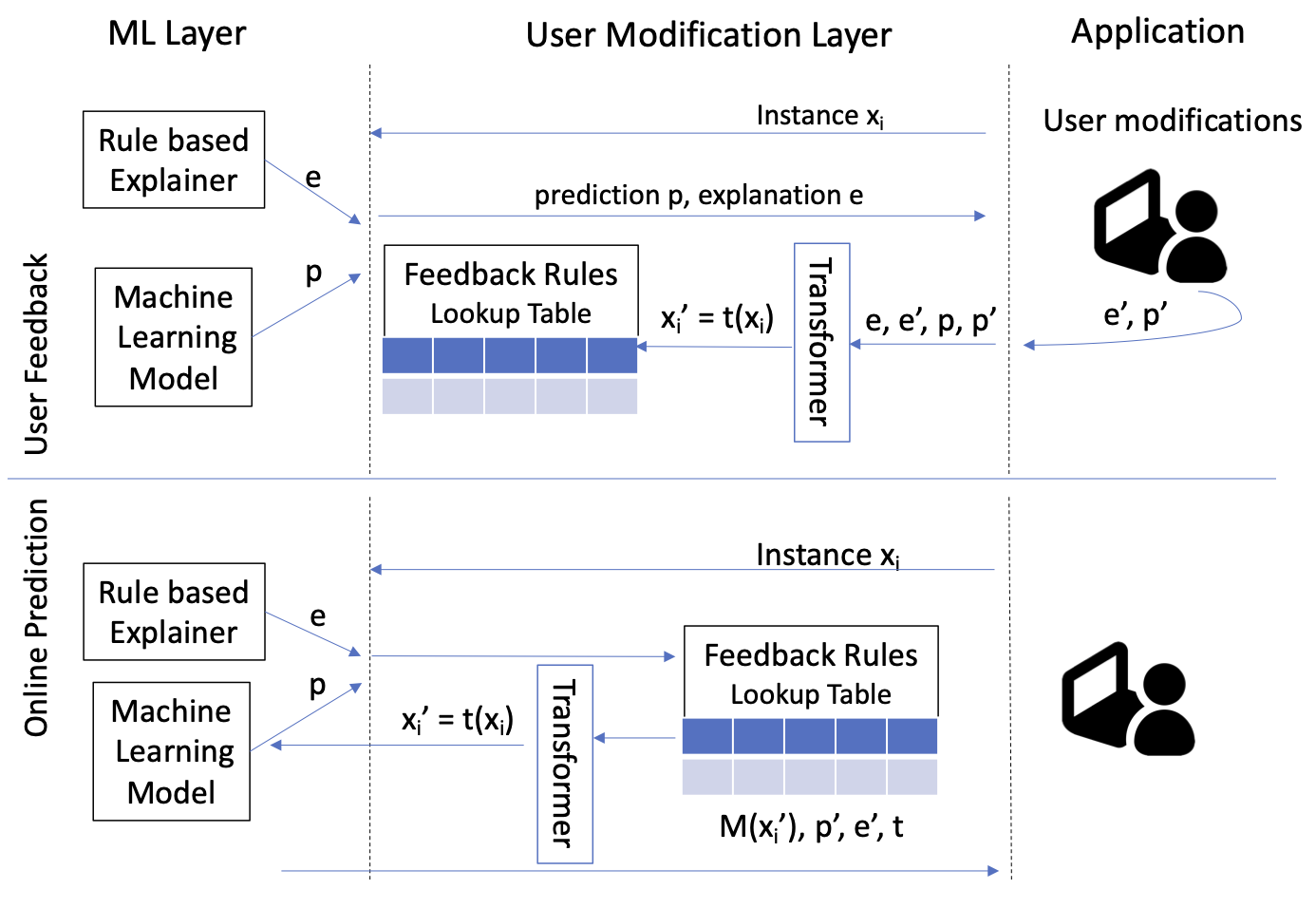}
\caption{Overview}
\label{fig:overview}
\end{figure}
The overall flow of our proposed interactive overlay solution is provided in Figure \ref{fig:overview}. The overlay includes three layers, namely, \textit{ML Layer, User Modification Layer and Application Layer}. To illustrate the framework the two main processes of the solution are also separated out in Figure \ref{fig:overview}: \textit{processing user feedback} (upper), and \textit{making online predictions} (lower). In order to provide feedback to the system, the user is presented with a response and given the option to make adjustments which are then stored in the Feedback Rules Lookup Table. When making online predictions, the solution generates a response by activating previous user changes through the Feedback Rules Lookup Table in order to influence the current prediction. 

The ML Layer contains the ML model and the Explainer. They are assumed to be provided to the interactive overlay. It should be emphasized that our solution is not dependant on any specific ML model. As can be seen from the online prediction phase, once the end-user sends an instance and asks for a response, this request passes directly to the ML Layer. 
The ML Layer then provides the initial prediction $p$ through calling the ML model along with an interpretable explanations $e$ which is provided by the Explainer [algorithm \ref{alg:GenerateResponse} line 2-3]. Assuming no relevant user feedback is found, the user is then presented with both the prediction and the explanation and given the opportunity to modify $e$ and/or $p$ [algorithm \ref{alg:GenerateResponse} line 13]. These modifications ($e'$ and $p'$) are then stored in the Feedback Rule Lookup Table along with an inferred transformation function $t$. When making an online prediction, the label $p$ is used together with the explanation $e$ for accessing the Feedback Rules Lookup Table [algorithm \ref{alg:GenerateResponse} line 5] and retrieving relevant feedback rules. 
The input instance $x_i$ is then evaluated against each one of the retrieved feedback rules [algorithm \ref{alg:EvaluateFeedbackRules} line 2-15]. The instance can either satisfy both, the original explanation rule and the feedback rule, or only one of the two rules [algorithm \ref{alg:EvaluateFeedbackRules} line 5]. The latter condition occurs when the instance falls in between the decision boundaries of the two rules. If the input instance $x_i$ does not satisfy the feedback rule, $p'$ is set to the other label [algorithm \ref{alg:EvaluateFeedbackRules} line 6-7]. Finally, if the model prediction $p$ does not match $p'$ the transformation is applied and the new result returned to the user [algorithm \ref{alg:EvaluateFeedbackRules} line 8-13].

\subsection{Explainer and Feedback Rules}
Our proposed solution can work with any ML model; however, since users can provide feedback on the explanations themselves, the solution requires both the explanations and feedback to have the same structure. With the increasing need for interpretability there have been a number of works recently that revisit Boolean rule set logic where the rules themselves are the model. This is particularly relevant for high stakes decision making such as in the medical domain or criminal prosecution \cite{rudin2019stop, angelino2017certifiably,rudin2018age}. These solutions are composed of \textit{if-then} statements that predict the label, and have the advantage that they are completely transparent and interpretable to the user. Drawing from these solutions, we assume that both the explanations and the feedback are in the form of an \textit{if-then statement}, or formally, \textit{Boolean rules}. In order to produce meaningful explanations to support the prediction from an ML model, we leverage the proposed BRCG framework \cite{dash2018boolean}. This framework produces a disjunctive normal form (DNF) representation of a logical formula to predict a class label for an instance, where the class label can be either the ground truth label or the label provided by an ML model. Although BRCG is proposed for binary classification models, it can be generalized to multi-class problems using a \textit{one-vs.-rest} configuration~\cite{bishop2006pattern}. Our solution can use other explainer models, as long as the explanations can be mapped to boolean rules. Given a binary classifier $M(Y|X)$ which predicts Y given X, two sets of rules are generated using BRCG which explain the predictions of the classifier for the two classes. These two rule sets together form the \textit{Explainer Rule Set (ERS)}, which is used by the ML Layer. The feedback given by the users is stored as a \textit{feedback rule} ($FR$). A \textit{feedback rule set (FRS)} is the set of feedback rules that are stored in Feedback Rules LookUp Table, as depicted in Figure \ref{fig:overview}.

\subsection{User Modification Layer}
The User Modification Layer is responsible for mapping user feedback into instance transformation functions. The user can modify a clause by adding or removing one or more conditions or modifying existing conditions by changing the comparison operator, the value or both. Given an explainer clause such as \(age>26 \land income>50k \land education=``Masters"\), examples of possible user modifications are: 
\begin{description}
    \item[add a condition] \(age>26\ \land income>50k\  \land education=``Masters" \land \textcolor{blue}{\pmb{occupation=``Sales"}}\)
    \item[delete a condition] \(\pmb{\stkout{\textcolor{blue}{age>26\ \land}}}\ income>50k\ \land education=``Masters"\)
    \item[modify a value] \(age>\textcolor{blue}{\pmb{30}}\ \land income>50k\ \land education=``Masters"\)
    \item[modify an operator] \(age\textcolor{blue}{\pmb{<}}26\ \land income>50k\ \land education=``Masters"\)
\end{description}


\begin{algorithm}[!t]
\small
\DontPrintSemicolon
 \KwInput{
 $x_i$: data input instance, \\ 
 \textit{ERS}: Explainer Rule Set, \\
 \textit{FRS}: Feedback Rule Set} 
  \KwOutput{\textbf{\textit{response}}: system response that applies to $x_i$}
    $response \gets \{\} $\\
    $p \gets QueryMlModel(x_i)$\\
    $explanations \gets explain(x_i, p, ERS)$\\
    \For{each $e$ in $explanations$} {
        $candidate\_fr \gets RetrieveFeedBackRule(e, p, FRS)$\\
        $response \gets \textbf{\textit{EvaluateFeedbackRules}}(x_i, e, p, candidate\_fr$)\\
    }
    \tcp{no feedback rule matching explanation $e$ and prediction $p$}
    \If{$response$ is empty}{
         \For{each $frs\_entry$ in $FRS$} {
            $e, candidate\_fr \gets frs\_entry$\\
            \If{$e$ not in $explanations$}{
                $response \gets$ \textbf{\textit{EvaluateFeedbackRules}}($x_i$, $e$, $p$, $candidate\_fr$)
            }
        }
    }
    \tcp{no feedback rule matching instance $x_i$}
    \If{$response$ is empty}{
        $response \gets p, \mathbf{random\_select}(explanations)$
    }
    return $response$
      
\caption{GenerateResponse}
\label{alg:GenerateResponse}
\end{algorithm}

\begin{algorithm}[!t]
\small
\DontPrintSemicolon
 \KwInput{
 $x_i$: data input instance, \\ 
 $e$: explanation rule, \\
 $p$: prediction, \\
 $candidate\_fr$: list of feedback rules to search for applicability to $x_i$ }
 \KwOutput{\textbf{\textit{response}}: system response that applies to $x_i$}
    $response \gets \{\} $\\
    \For{each $fr$ in $candidate\_fr$} {
        $R', t \gets fr$ \\
        $e', p' \gets R'$ \\
        \If{$x_i$ satisfies $e$ \textbf{or} $x_i$ satisfies $e'$}{
            \If{$x_i$ not satisfies $e'$} {$p' \gets GetOtherLabel(p')$} 
            \If{ $p \neq p'$}{
                $x_i' \gets t(x_i)$ \\
                $new\_p \gets QueryMlModel(x_i')$ \\
                $response \gets new\_p, p', e', t.description$\\
                \If{$new\_p = p'$}{
                    return $response$
                }
            }
            \Else{
                \tcp{model is already capturing this rule} 
                return $response \gets p, p', e'$ 
            }
        }
    }
    return $response$
\caption{EvaluateFeedbackRules}
\label{alg:EvaluateFeedbackRules}
\end{algorithm}  


\subsection{Instance Transformation}\label{subsec:instance transformation}

Given a prediction $p$ from a binary classifier $M(Y|X)$ and an explanation clause $e$, there can be 3 possible outcomes~\cite{teso2019explanatory}: 1) the prediction is correct and the explanation is correct as well (\textit{right for the right reasons}). 2) the prediction is correct but the explanation is not (\textit{right for the wrong reasons}). 3) the prediction is wrong and the explanation is also necessarily wrong (\textit{wrong for the wrong reasons}). Our research focuses on enabling the user to provide feedback in order to correct the model in cases 2 and 3. Whenever the prediction is correct, but the user is not satisfied with the explanation clause, they can modify the clause to correctly reflect the reason behind the prediction (case 2). More formally, given an explanation clause $e$ and a user input correction clause $e'$, we define a function $t: X \to X$ such that Equation \ref{eq:exp1} holds.
\begin{equation}
\begin{split}
    \forall i \in \{1,...,n\}, e'(x_{i}) = 1 \iff e(t(x_{i})) = 1
\end{split}
\label{eq:exp1}
\end{equation}
Intuitively, the transformation $t$ modifies only the instances that fall between the boundaries defined by $e$ and $e'$. The instances are pushed inside or pulled outside the decision boundaries of the ML model for the target class $p$ depending on whether the user is relaxing or narrowing the original boundaries defined by $e$. For instance, given as explanation $e$ the clause \(age>26 \land income>50k \land education=``Masters"\) with a predicted class label $p = 1$, and a user input correction clause $e'$ as \(age>\mathbf{30} \land income>50k \land education=``Masters"\), the function $t$ is defined as \texttt{if ($age > 26 \land age \leq 30$) then $age = 25$}. As the clause $e'$ narrowed the boundaries on the feature age, the transformation is pushing the instances that fall within the interval defined by $e$ ($age > 26$) and $e'$ ($age \leq 30$) outside the decision boundaries of the ML model for class $p = 1$. The \textit{margin} between the new value assigned to the numeric feature and the boundaries defined by $e$ and $e'$ is configurable per feature and by default is set to $1$.
For a categorical feature the transformation $t$ draws a new value from the feature domain. Given the explanation clause $e$ presented above and a correction clause $e'$ as \(age>26 \land income>50k \land education=\mathbf{``Doctorate"}\) where the condition on the categorical feature \textit{education} has been changed, the inferred transformation $t$ is described in Algorithm~\ref{alg:TranfWithCatF}.

\begin{algorithm}
\small
\DontPrintSemicolon
 \KwInput{
 \textit{$x_i$}: data input instance, \\ 
 \textit{E}: set representing the domain of feature education \\
 }
 \KwOutput{\textbf{$x_i'$}: transformed instance}
    $value \gets x_i.education$ \\
    \If{value = ``Doctorate"}{
        $new\_value \gets ``Masters"$ \\
    }
    \ElseIf{value = ``Masters"}{
        $new\_value \gets$ \textbf{ \textit{random\_select}}($E - \{``Masters"\}$)\\
    }
    $x_i.education \gets new\_value$ \\
    
    return $x_i$\\
\caption{Example of a transformation on a categorical feature when the class label is preserved}
\label{alg:TranfWithCatF}
\end{algorithm}  

The user is also able to correct a model when a prediction is wrong (case 3) by changing the label. Formally, given a clause $e$, a prediction label $p$, a user input correction clause $e'$ and a user label $p'$ where $p'\neq p$, the transformation function $t: X \to X$ is defined such that Equation \ref{eq:exp2} holds.
\begin{equation}
\begin{split}
    \forall i \in \{1,...,n\}, e'(x_{i}) = 1 \iff e(t(x_{i})) = 0
\end{split}
\label{eq:exp2}
\end{equation}
Intuitively, the transformation $t$ pushes all instances outside the decision boundaries of the ML model for class label $p$. For example, given as explanation $e$ the clause \(age>26 \land income>50k \land education=``Masters"\), a user correction clause $e'$ as \(age>\mathbf{30} \land income>50k \land education=``Masters"\), and a new predicted class label $p'$ such that $p \neq p'$, the resulting transformation $t$ is shown in Algorithm~\ref{alg:TranfWithNumF}. For pseudo-code documenting each variation of the transformation algorithms please see the supplementary material.

\begin{algorithm}
\small
\DontPrintSemicolon
 \KwInput{
 \textit{$x_i$}: data input instance, \\ 
 \textit{margin}: margin between the new value and the boundaries
 }
 \KwOutput{\textbf{\textit{$x_i'$}}: transformed instance}
    $value \gets x_i.age$ \\
    
    \If{value $>$ 30}{
        $new\_value \gets 26 - margin$ \\
    }
    \ElseIf{value $\leq$ 26}{
        $new\_value \gets 26 + margin$ \\
    }
    $x_i.age \gets new\_value$ \\
    
    return $x_i$\\
\caption{Example of a transformation on a numeric feature when the class label is changed}
\label{alg:TranfWithNumF}
\end{algorithm}  


It is important to note that for the purposes of this paper we assume an approximation for the decision boundaries are given by the clauses of the BRCG framework and so the instance transformation uses these decision boundaries to transform the input instance. Other options to consider could be leveraging prior work on detecting `flip points' for these values \cite{yousefzadeh2020auditing}.

\begin{figure*}[!t]
\captionsetup[subfigure]{justification=centering}
\centering
\subfloat[Tic-Tac-Toe]{
\includegraphics[ width=.5\columnwidth]{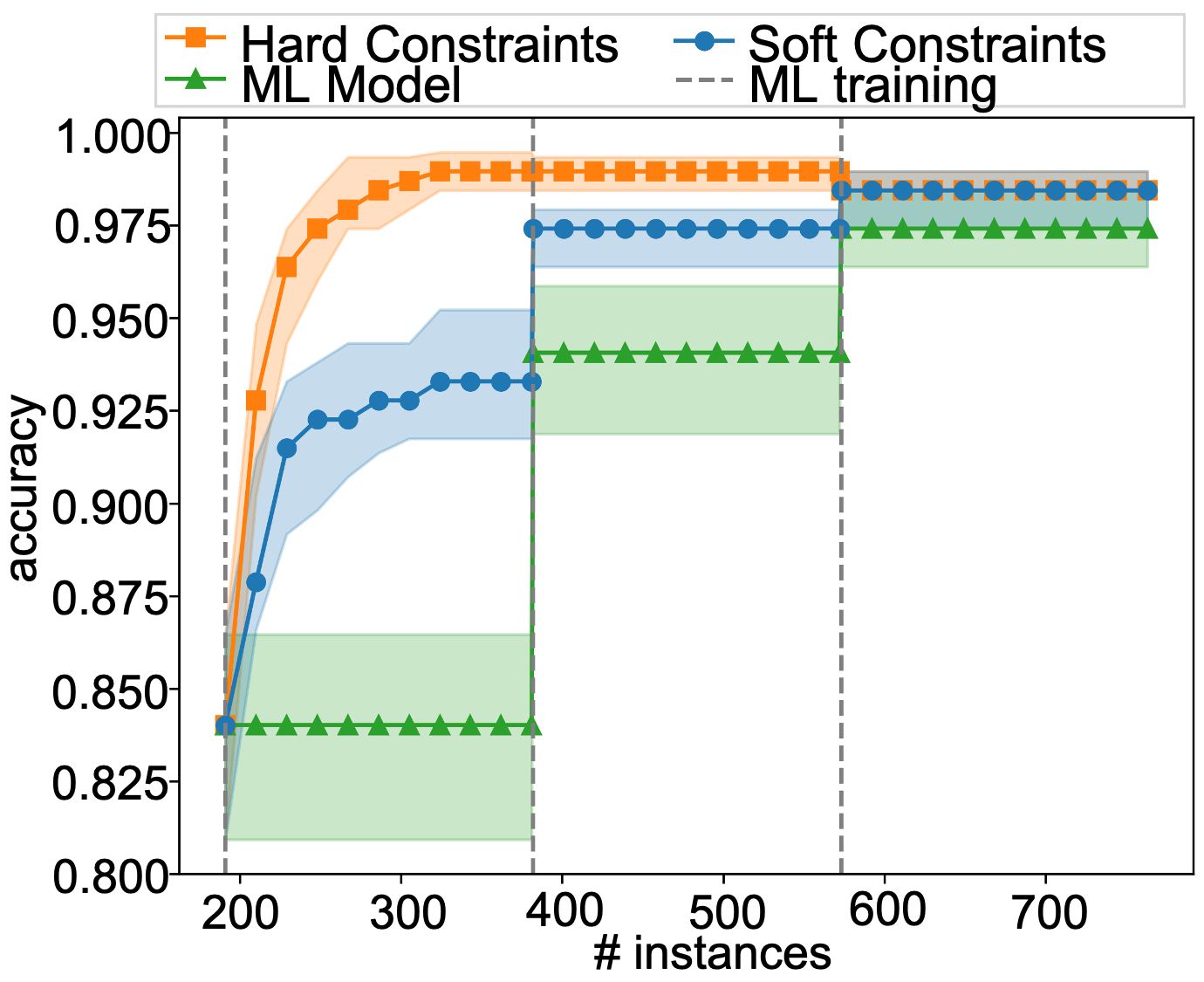}
}
\subfloat[Bank-mkt ]{
\includegraphics[width=.5\columnwidth]{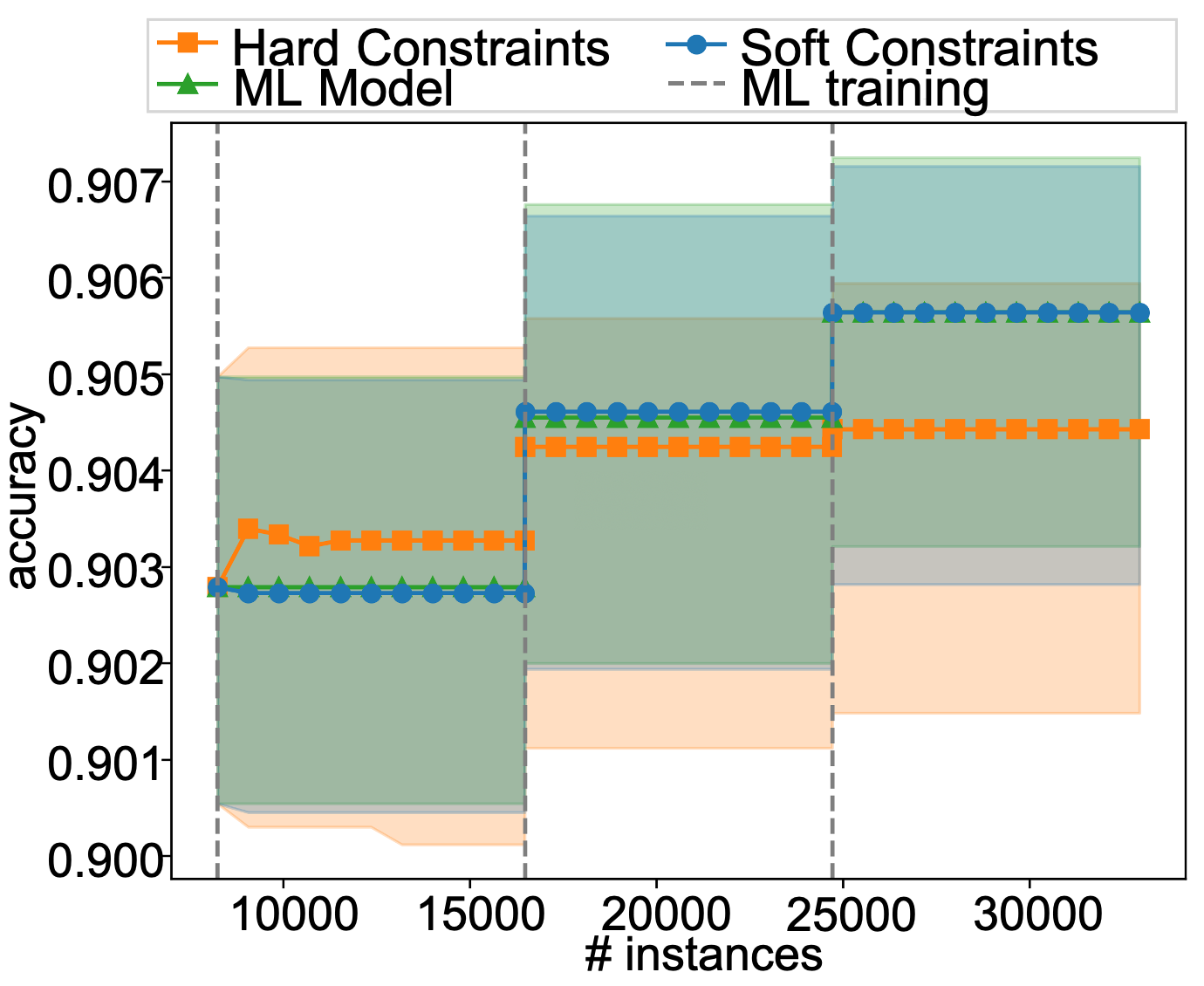}
}
\subfloat[Breast Cancer]{
\includegraphics[width=.5\columnwidth]{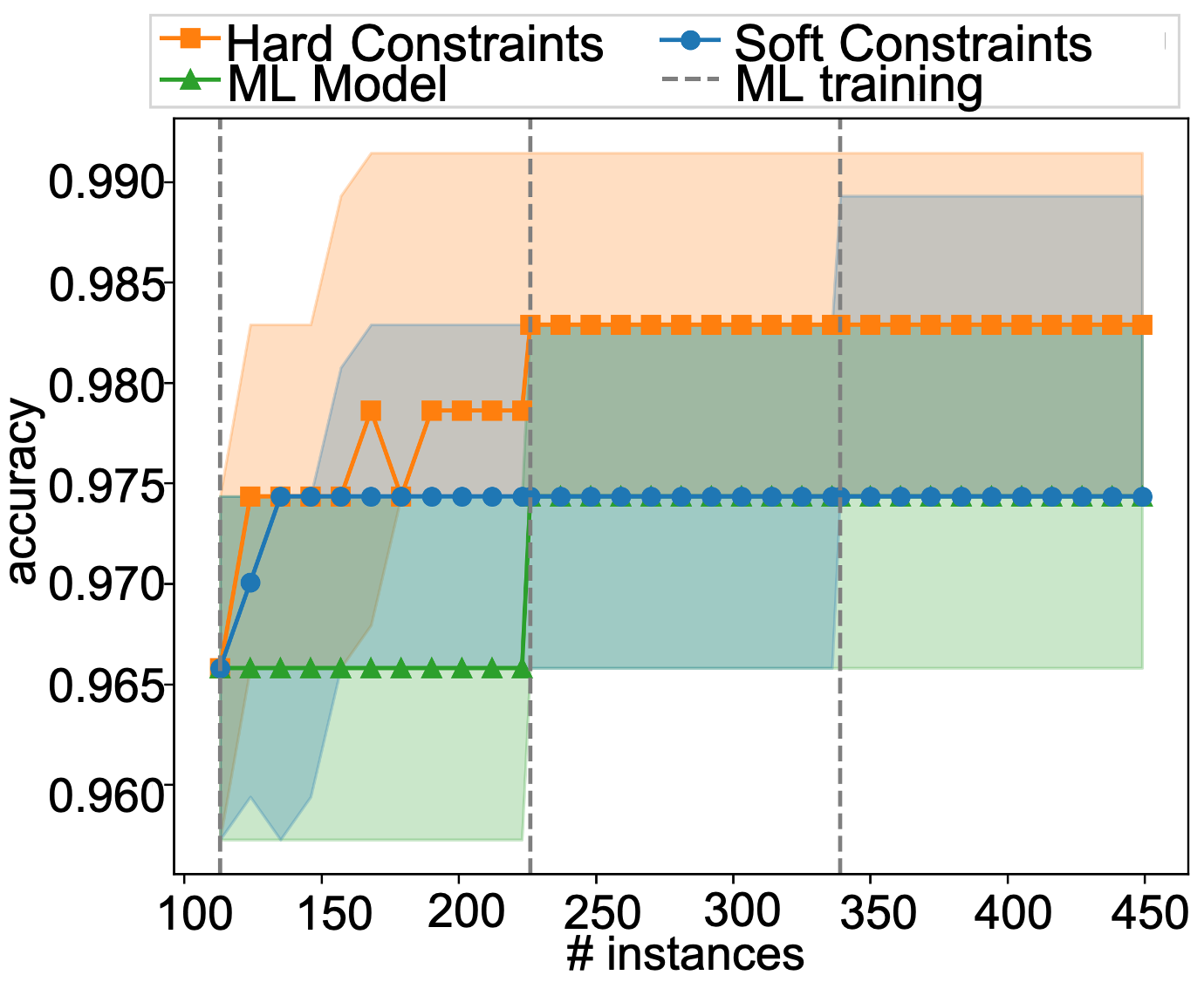}
}
\subfloat[Banknote ]{
\includegraphics[width=.5\columnwidth]{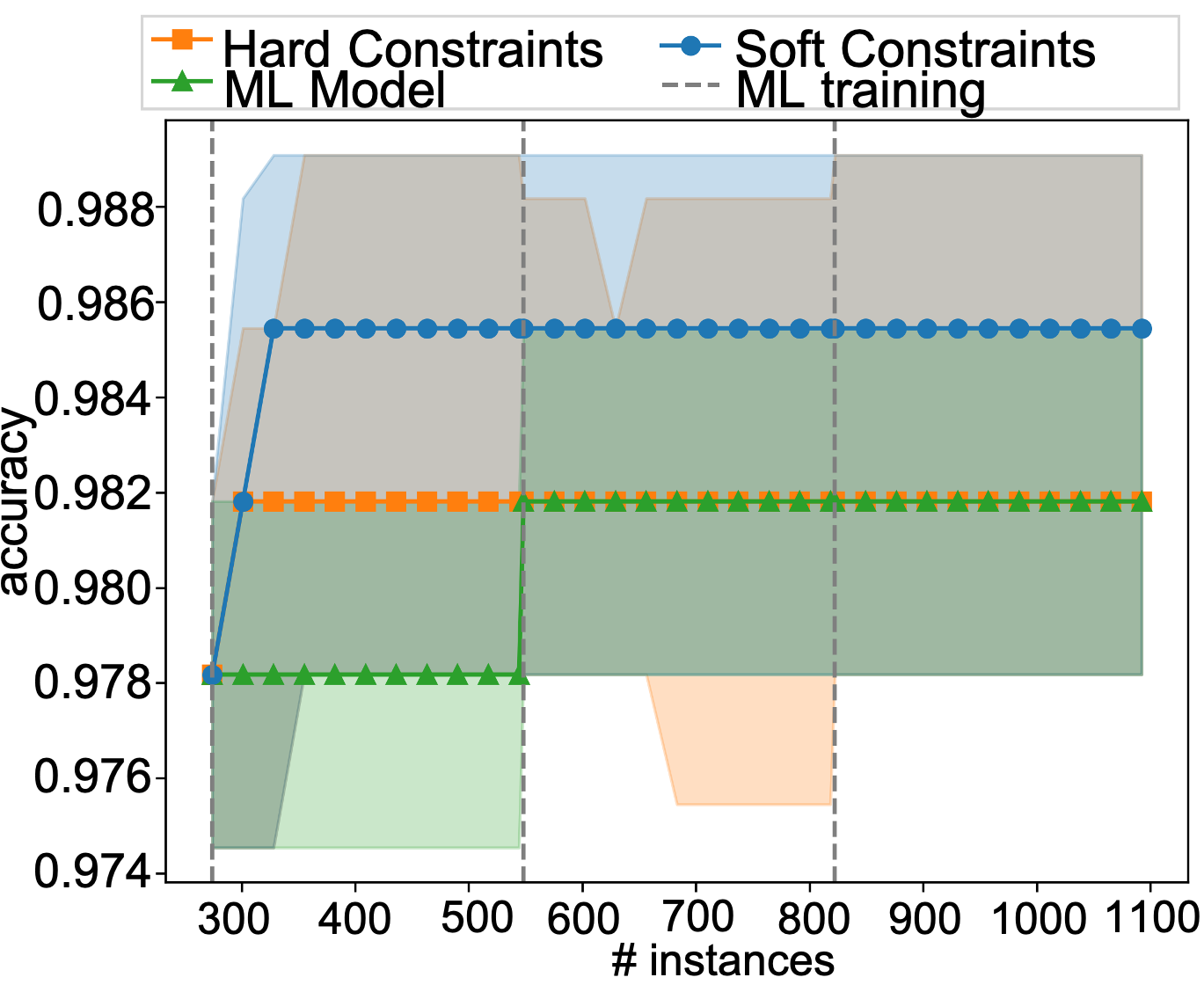}
}
\caption{Experiment 1 - Demonstrating Interactive Overlay Approach (Median and 25-75 percentiles)}
\label{fig:eval1}
\end{figure*}

\begin{figure*}[t]
\captionsetup[subfigure]{justification=centering}
\centering
\subfloat[Tic-Tac-Toe]{
\includegraphics[width=.5\columnwidth]{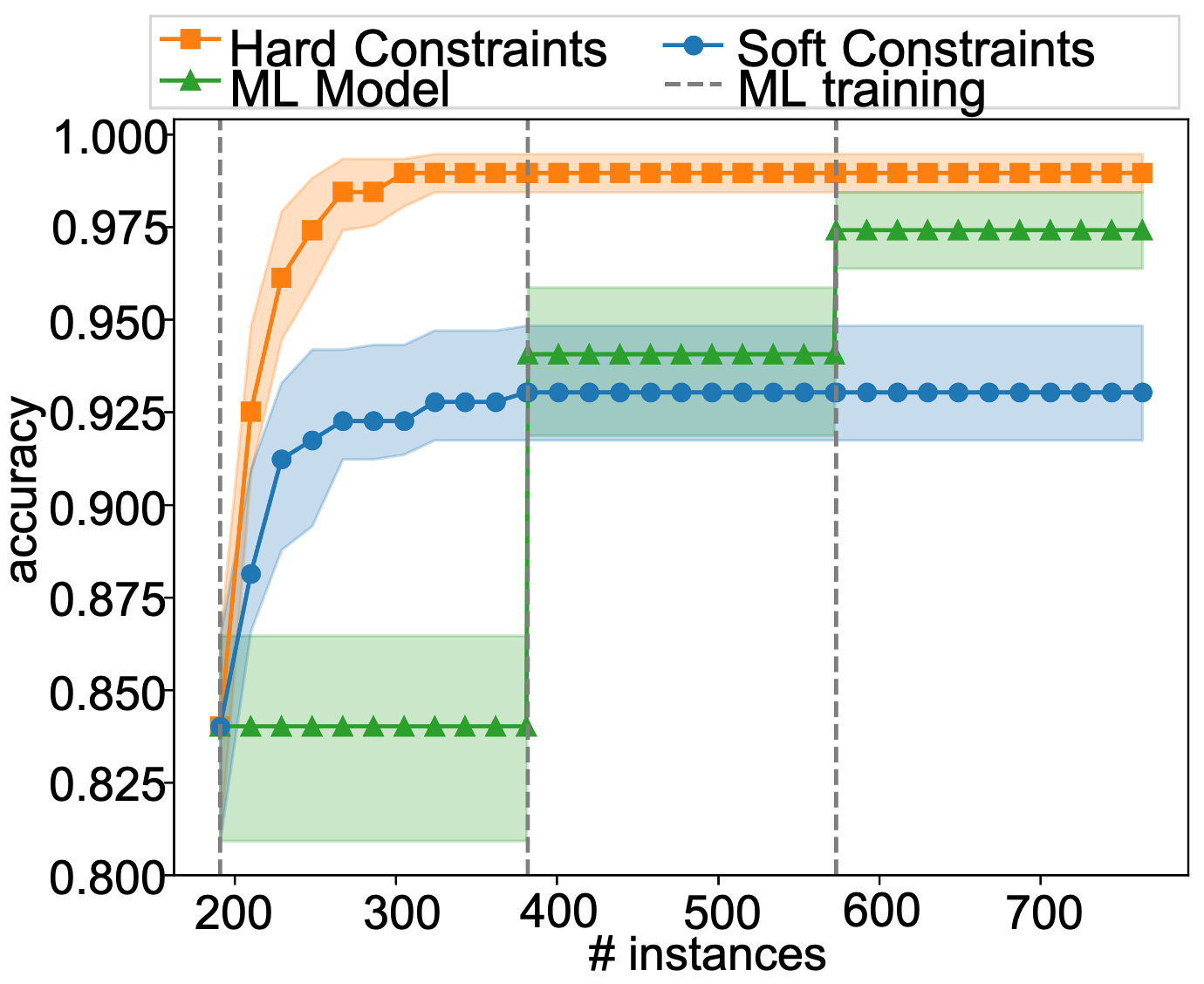}
}
\subfloat[Bank-mkt]{
\includegraphics[width=.5\columnwidth]{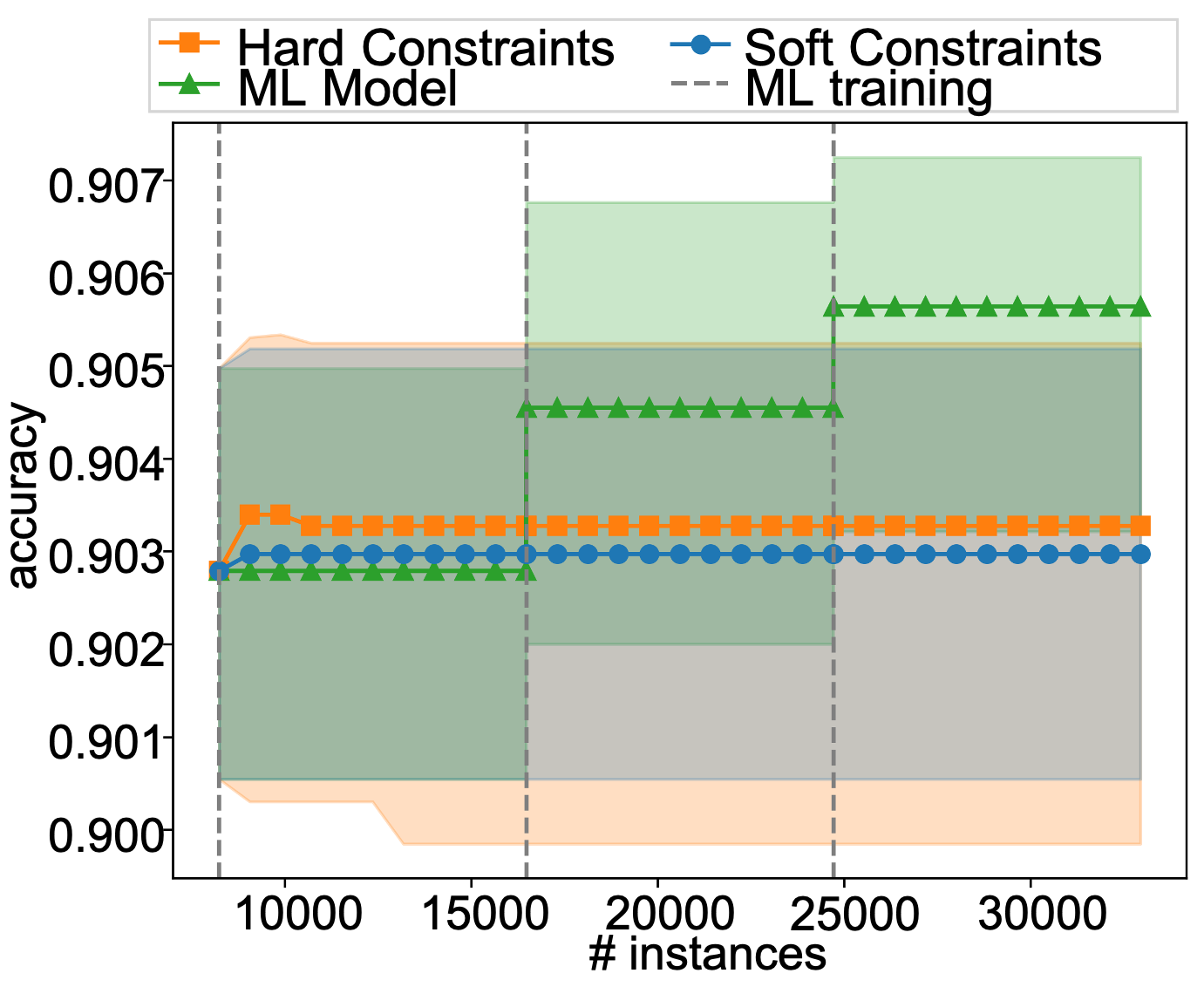}
}
\subfloat[Breast Cancer]{
\includegraphics[width=.5\columnwidth]{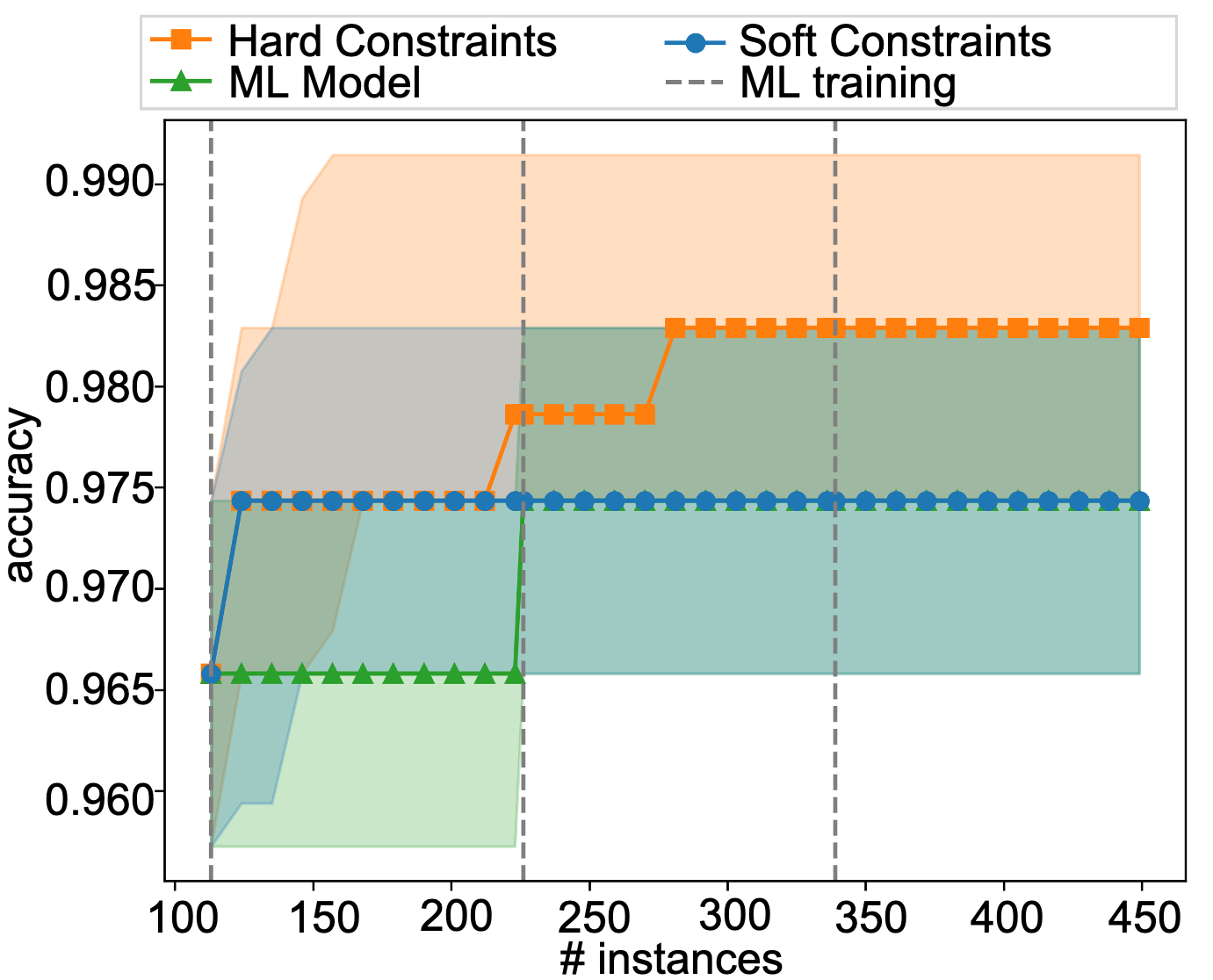}
}
\subfloat[Banknote]{
\includegraphics[width=.5\columnwidth]{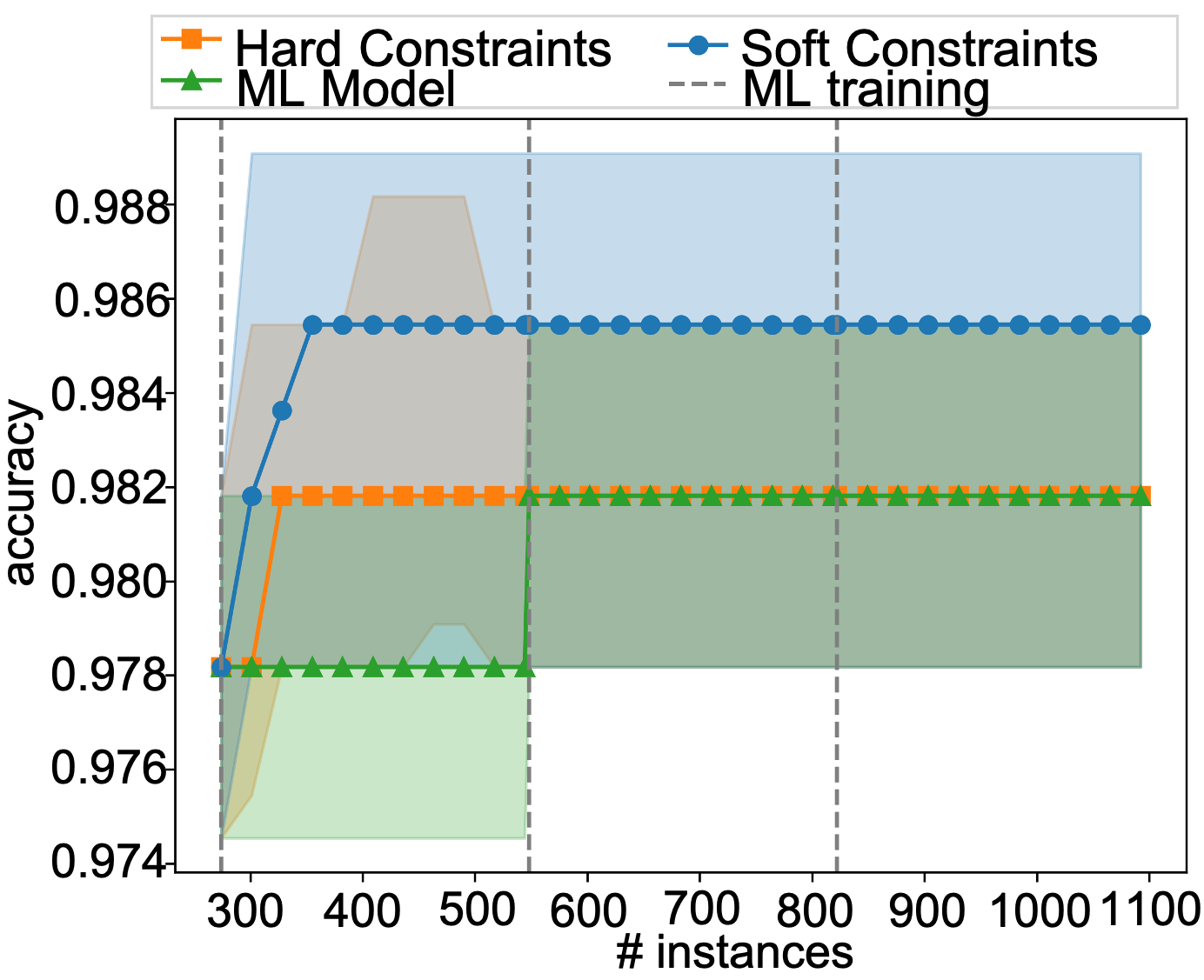}
}
\caption{Experiment 2 - Demonstrating Interactive Overlay Approach with underlying ML model only trained at 20\% }
\label{fig:eval2}
\end{figure*}


\subsection{Response Generation}
Figure \ref{fig:overview} shows that when an input instance $x_i$ is presented to the system, the prediction label $p$ and the explanation $e$ are used as a lookup to the Feedback Rules Lookup Table in order to retrieve any Feedback Rules $FR$ [algorithm \ref{alg:GenerateResponse} line 5]. Given our rules are not necessarily non-overlapping, it could be possible that there are multiple explanations for a single instance, in this case we evaluate all explanations. 
When an instance $x_i$ satisfies at least one of the two rules $e$ and $e'$, and $p'$ differs from the model prediction $p$, then the transformation function $t$ is applied to generate $x_i'$ which is then given as new input to the machine learning model. If the transformation results in a modification of $p$ to $p'$ then the modified result is returned to the user along with the user contributed modification $e'$ and the transformation performed [algorithm \ref{alg:EvaluateFeedbackRules} line 8-13]. If no user rule results in a modification we return the last seen $FR$ that applied to the instance or whose related explanation $e$ applied to the instance. If no user rules are found at all we return the explanation rule $R$ to the user with the option for the user to provide feedback. 

A feedback rule may have decision boundaries that are looser than those defined by the original explanation rule, that is, an input instance that satisfies the feedback rule may not satisfy the original explanation rule. To correctly handle this scenario all remaining feedback rules are evaluated against the instance [algorithm \ref{alg:GenerateResponse} lines 8-11]. One challenge that exists is that rule based explanations may change after model retraining and as a result there may be rules in the Feedback Rule Lookup Table that are no longer returned from the Explainer; however, the system will still need to honour them. The evaluation of all remaining feedback rules accounts for properly handling this scenario as well.

As some domains may require \textbf{\textit{hard constraints (hc)}} such as regulatory compliance, we return both the transformed prediction $M(x_i')$ as well as the user input prediction $p'$. If the application requires the constraint to be treated as a hard constraint then $p'$ can be used, if the user feedback is treated as a \textbf{\textit{soft constraint (sc)}} then the prediction on the transformed instance is used. 

\section{Experimental Evaluation}
In order to address our research questions $R1$ and $R2$ we want to evaluate if we can improve the performance of an underlying machine learning model in between retrains by combining the ML predictions with user contributed modifications to prediction explanations. 

\subsection{Methodology}
To mimic the contributions of a user we employ an oracle based approach where the rule based explainer is trained on 100\% of the data, we call this Full Knowledge Rule Set ($FKRS$). This approach is similar to mechanisms used to evaluate active learning approaches \cite{kulkarni2018interactive}. For the purposes of these experiments the underlying machine learning algorithm used is a logistic regression with 500 iteration limit\footnote{For our 4 test datasets the logistic regression model reaches its max accuracy in less iterations.}. Numeric features are pre-processed with StandardScaler and the categorical one with a OneHotEncoder. We provide results for both the \textit{hard constraint} (\textsc{hc}) approach of our solution and the \textit{soft constraint} (\textsc{sc}) approach. Our accuracy measure is the ratio of the number of correct predictions to the total number of samples. We select four well known binary classification benchmarks from the UCI repository\footnote{\url{https://archive.ics.uci.edu/ml/datasets/}} \textsc{tic-tac-toe}, \textsc{banknote}, \textsc{bank-mkt} and \textsc{breast cancer}.

\begin{figure*}[t]
\captionsetup[subfigure]{justification=centering}
\centering
\subfloat[Tic-Tac-Toe]{
\includegraphics[width=.5\columnwidth]{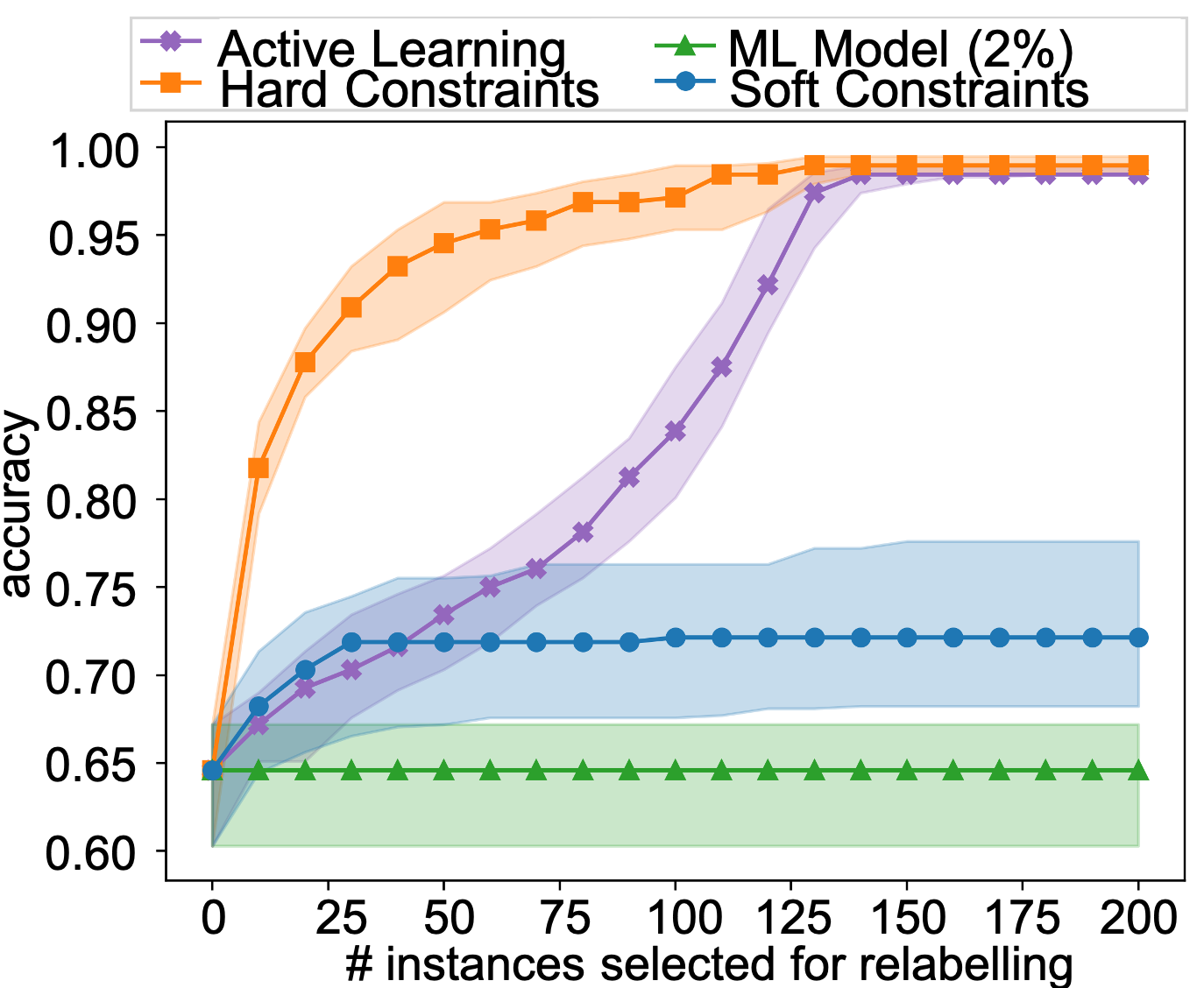}
}
\subfloat[Bank-mkt]{
\includegraphics[width=.5\columnwidth]{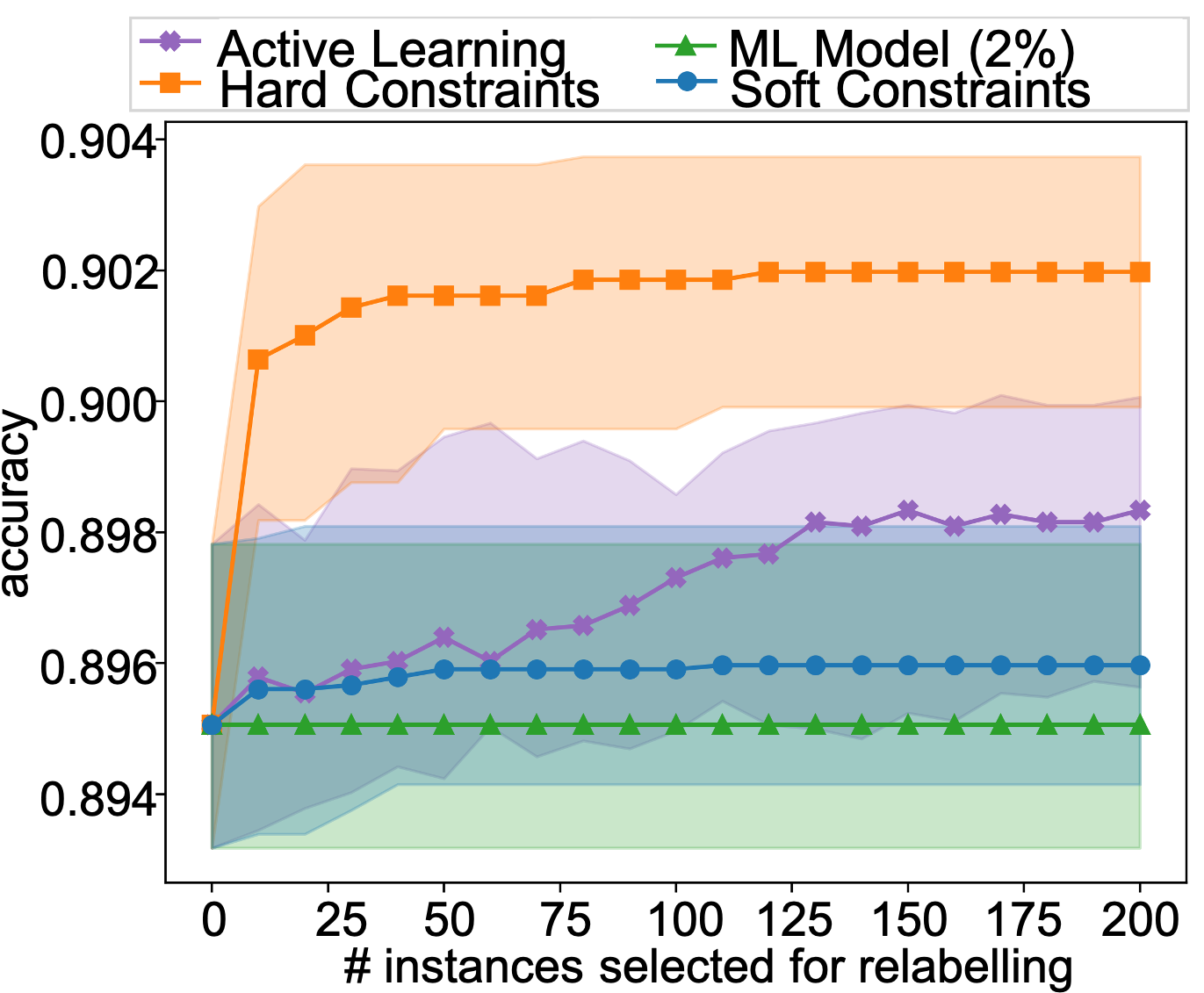}
}
\subfloat[Breast Cancer]{
\includegraphics[width=.5\columnwidth]{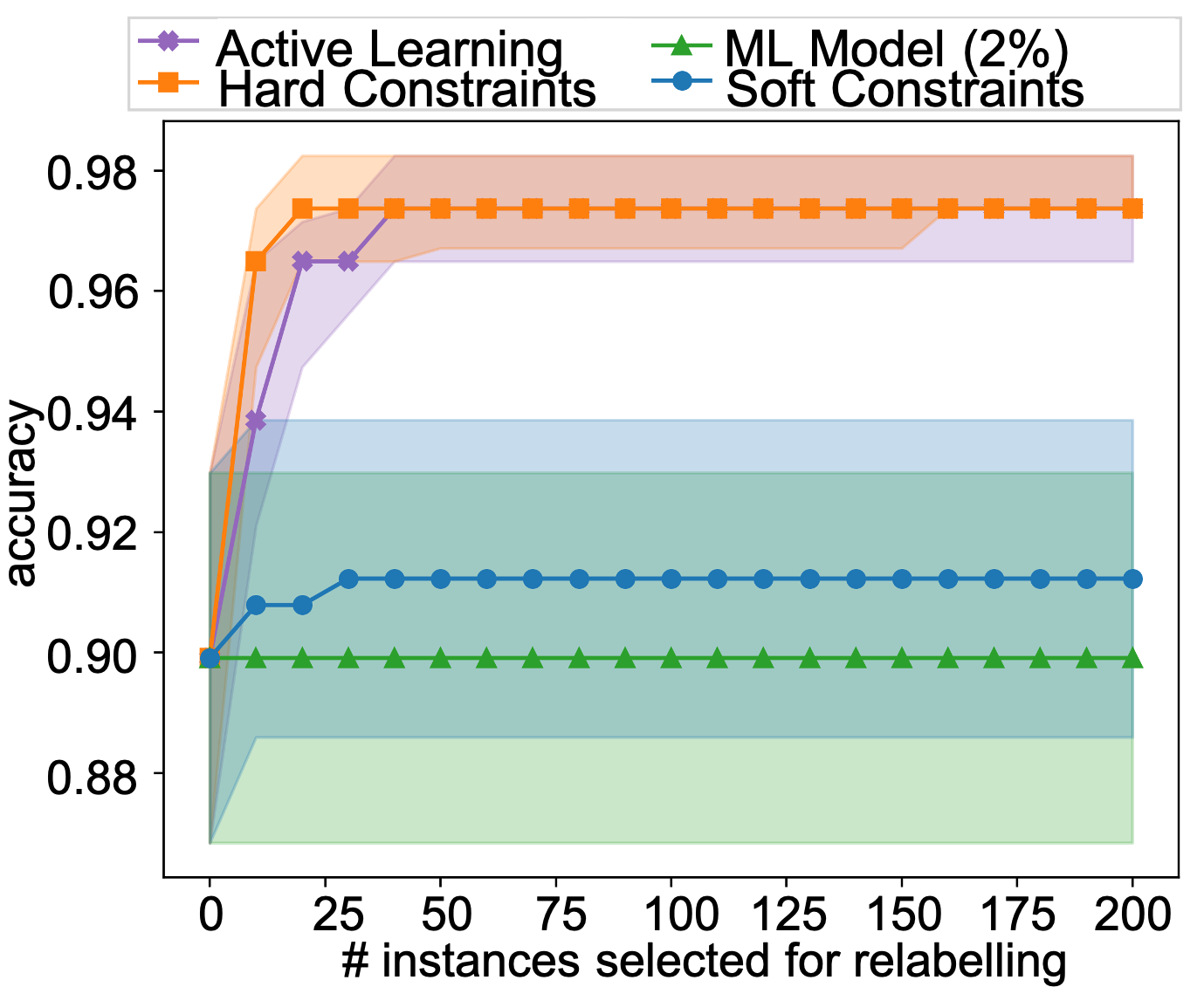}
}
\subfloat[Banknote]{
\includegraphics[width=.5\columnwidth]{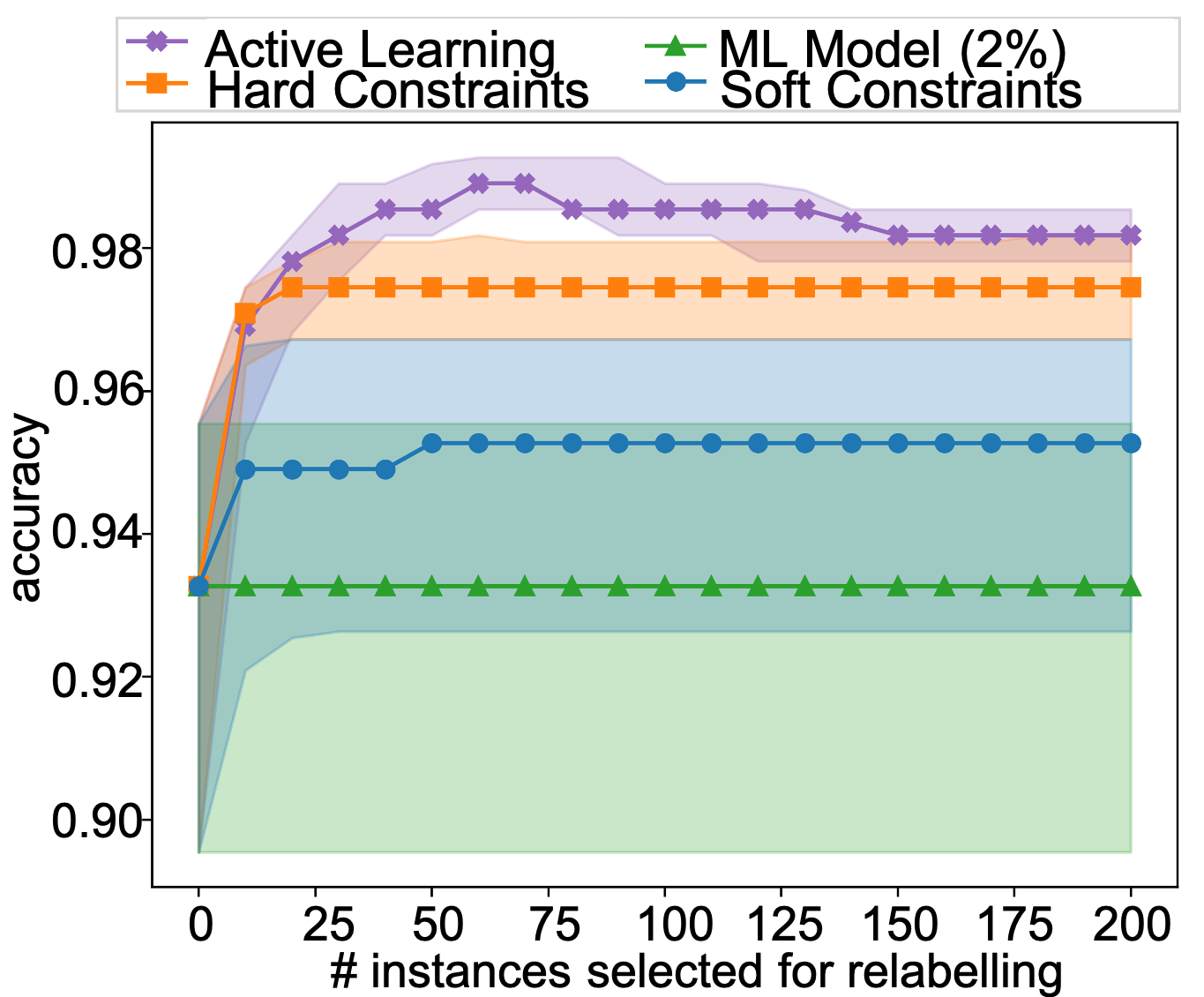}
}
\caption{Experiment 3 - Demonstrating Interactive Overlay Approach vs Active Learning. ML model is trained at 2\% and retrained on each batch of 10 instances selected for Active Learning whereas it is never retrained for the Overlay solution}
\label{fig:eval3}
\end{figure*}

\textbf{Experiment 1:} 
To evaluate R1 we assume both the overlay solution and the ML algorithm learn on the same data. However, we assume user corrections may be provided to the overlay solution from the oracle $FKRS$. \textbf{Train:} The data is divided into 80\% for training and 20\% for the hold-out test set. The training data is further divided into four slices representing 20\% of the data. The ML model is trained on the first 20\% and the resulting model is used to train the rule based explainer learnt on the labels provided by the ML model, we refer to the resulting ruleset as the Partial Knowledge Rule Set ($PKRS$). \textbf{Online Learning:} The evaluation presents each data instance of the next slice to the overlay solution. Whenever the overlay solution does not find a user feedback rule that applies to a data instance, the explanation clause and prediction label provided by $PKRS$ are compared with those output by the oracle and in the cases where they differ, the oracle output is added as a user feedback rule to the lookup table of the overlay solution. After each slice has been processed the slice is added to the training data and the ML model is retrained. The $PKRS$ is also retrained on the updated ML model. Given we aim to improve performance between model retrains, we calculate the accuracy of the overlay solution every additional $10\%$ of data slice, whereas the ML accuracy is only recomputed after each additional retrain. We repeat the experiment $50$ times, each time using a different shuffle of the whole dataset, therefore producing different slices for training and testing each time. (For the evaluation simulation algorithm please see the Supplemental material).

\textbf{Experiment 2:} To evaluate R2 where we need to assess whether the overlay solution can achieve comparable performance when given access to less training data, we repeat experiment 1 above with the exception that the underlying ML model used by the overlay solution is not retrained after the initial training on 20\%, whereas the pure ML based solution is retrained on the initial 20\% data plus each processed slice.

\textbf{Experiment 3:} We compare our solution to an Active Learning approach. 
The optimal conditions for comparing the approaches is with an ML model whose performance has room for improvement. Hence, for this experiment we trained the model so that its initial performance is relatively low similar to related work~\cite{ghai2020explainable}. We use 20\% of the data as hold out set for evaluation and we initially trained the ML model on 2\% of the data while the remaining 78\% of the data is used as a data pool from which the active learning selector draws the instances to use for retraining.
We adopted a low margin strategy~\cite{scheffer2001active} to select the instances to provide to the oracle for labelling. At each iteration a batch of 10 instances is selected from the pool, labelled by the oracle with the ground truth, and then used for retraining the ML model. We repeat these steps 20 times and after each retraining the accuracy of the model is measured against the test set. Our overlay solution is also initialized with a ML model trained on 2\% of the data, but this model is never retrained. The batch of instances drawn by the selector are used to test the predictions of the framework and provide corrections in the form of ground truth labels and explanation clauses from $FKRS$. After a batch is processed, the accuracy of the overlay solution is measured against the testing set. As with the previous experiments it is repeated $50$ times each time using a different shuffle of the whole data.

We implement all algorithms in Python using scikit-learn~\cite{pedregosa2011scikit} and perform the experiments on a cluster of Intel Xeon CPU E5-2683 processors at 2.00GHz with 8 Cores and 64GB of RAM. For the BRCG implementation we use the "Light" version of BRCG~\cite{arya2019one} available in the open source library AIX360\footnote{\url{https://github.com/Trusted-AI/AIX360}}. Table \ref{tab:table1} shows the details of the dataset used along with the overall accuracy of the BRCG solution in generating boolean rules that reflect the decision processes. As can be seen, the accuracy varies across datasets; however, for the purposes of this paper the goal is not to have a complete rule based system that delivers 100\% accuracy but to provide clauses rich enough for a user to contribute modifications and demonstrate that the final predictions reflect those modifications. Table~\ref{tab:table2} shows an example of DNF rules generated during our experiments, specifically the one that represents the Full Knowledge Rule Set for the Bank Marketing dataset.

\begin{table}
    \small
   
    \begin{tabular}{|l|l|l|l|l|} 
    \hline \hline
    & tic-tac-toe   &  bank-mkt & b-cancer  & banknote\\
    \hline
    \# instances & 958           &  45,211   & 569           & 1,372 \\
     \hline
    \# features    & 9             &   17      & 32            & 5\\
    \hline
   BRCG Acc         & 0.992         &   0.903   & 0.976         & 0.976\\
       \hline \hline
    \end{tabular}
    \caption{Datasets}
    \label{tab:table1}
\end{table}

\begin{table*}
    \small
 
    \begin{center}
    \begin{tabular}{|l|l|} 
    \hline \hline
    Class & DNF rule \\
    \hline
    no & \((nr.employed > 5076.20) \lor (poutcome \neq ``success" \land duration \leq 368.00)\) \\
     \hline
    yes & \((duration > 280.00 \land emp.var.rate \leq -3.00) \lor (poutcome = ``success" \land nr.employed \leq 5076.20)\) \\
     \hline \hline
    \end{tabular}
    \end{center}
   
    \caption{DNF rules generated with the Light BRCG method for the Bank Marketing dataset. 
    The classification goal is to predict if the client will subscribe (yes/no) a term deposit. According to the UCI data dictionary, \emph{nr.employed} is the number of employees, \emph{poutcome} is the outcome of the previous marketing campaign, \emph{duration} is the last contact duration, in seconds and \emph{emp.var.rate} represents the employment variation rate.}
     \label{tab:table2}
\end{table*}

\subsection{Results}
Figure \ref{fig:eval1} shows the results for Experiment 1. As can be seen, the \textsc{sc} approach is able to learn user modifications that improve the baseline ML performance between model retrains, meaning we can support online changes without compromising accuracy. In most cases the \textsc{sc} approach performs better than the \textsc{hc} approach, demonstrating that the rule based approach still benefits from the underlying predictive value of the ML model. Interestingly, for the \textsc{tic-tac-toe} dataset the \textsc{hc} outperforms both the ML and \textsc{sc} approach. This is due to the deterministic nature of the \textsc{tic-tac-toe} domain which is by definition rules based. 

Figure \ref{fig:eval2} shows the results for Experiment 2 where the ML model is retrained after each additional 20\% of the data but the ML model leveraged by the \textsc{sc} approach is kept at 20\% so any improvement seen is through simulated user feedback rules. The impact is clearly seen on the performance in the \textsc{tic-tac-toe} dataset where the ML model outperforms the \textsc{sc} after the first retrain. For the \textsc{banknote} data, on the other hand, we see relatively little impact on performance, demonstrating enough information was captured in the first 20\% of the data when combined with user modifications.  

As can be seen from figure \ref{fig:eval3}, the \textsc{sc} based approach is quickly overtaken by the active learning approach where the model is retrained. This is unsurprising given the ML model leveraged by \textsc{sc} is only trained on 2\% of the data and does not benefit from retraining. What is interesting, however, is that the \textsc{hc} approach rapidly outperforms the active learning approach, demonstrating that feedback in rule format can have a higher impact on performance than simple labelling. The quick divergence between \textsc{sc} and \textsc{hc} could potentially be used as a signal to the solution that the underlying ML model is no longer sufficient to capture the current decision processes and trigger a model retrain only when needed.

Our experiments above have demonstrated the ability to support model adjustments to an already trained model. Given the need for user input, our solution most benefits from adjustments to interpretable features. As seen from the tic-tac-toe case, when it is possible to express rules that clearly reflect the underlying decision making process then the overlay can improve solution accuracy. One limitation of soft constraints is seen when the ML model has far too little knowledge and the rule reflected in the model may be too different from the target rule, in which case the hard constraint logic becomes necessary. As a result, we believe this framework is most useful when making adjustments and corrections to existing rules related to interpretable variables e.g. age threshold or adding additional categorical values.

\subsection{Conclusion}

Given the reliance on AI for important decision making problems, the area of explainability has become a crucial subject of research. Advances in explainability have made it possible to consider opportunities in Interactive Machine Learning (IML) which seek to include the domain experts in the model creation process by allowing them to give feedback, explore the impact and tune their feedback accordingly \cite{amershi2014power}. In this paper we have presented a solution that allows users to provide modifications to interpretable boolean rules. We have demonstrated that these rules can be applied in an online fashion in order to cope with dynamic policy changes. Machine learning models in regulated enterprises, such as finance and healthcare, struggle to encode policy directives because these directives are generally not part of the data fabric. Our approach invites policy experts to encode domain-specific directives and applies appropriate transformations to existing ML decision boundaries at the time they take effect. Once the ML training procedure has caught up with the policy change the decision rule could be promoted from the interaction layer to a constraint in the ML training process \cite{thomas2019preventing}.

\bibliography{references}

\end{document}


\linenumbers

\maketitle

\begin{abstract}
Here, we provide a supplementary material for "User Driven Model Adjustment via Boolean Rule Explanations" (Paper Id: 7173). 
\end{abstract}
\section{Evaluation Supplementary Material}

\subsection{Rule Modification examples}
In this subsection we present some of the rule modifications used in the experiments.\\
From the Bank Marketing dataset:\\
partial knowledge rule: $poutcome \neq ``success'' \land duration \leq 368.00$; class NO\\
user corrected rule: $poutcome \neq ``success'' duration \leq 548.00$; class NO\\
modifications: changed duration value from $368.00$ to $548.00$.
\\
From Tic-Tac-Toe dataset:\\
partial knowledge rule: $tr \eq ``x'' \land br \neq ``o'' \land bl \neq ``o''$; class x-wins\\
user corrected rule: $tr \eq ``x'' \land mr \eq ``x'' \land br \eq ``x''$; class x-wins\\
modifications: changed $br \neq ``o''$ to $br \eq ``x''$; deleted $bl \neq ``o''$; added $mr \eq ``x''$.
\\
From Breast-cancer dataset:\\
partial knowledge rule: $texture\-worst > 20.08 \land smoothness\-mean > 0.10 \land concave points\-worst > 0.15$; class M\\
user corrected rule: $texture\-worst > 23.58 \land symmetry-worst > 0.26 \land concave points-worst > 0.15$; class M\\ 
modifications: changed texture-worst value from $20.08$ to $23.58$; deleted $smoothness\-mean > 0.10$; added $symmetry\-worst > 0.26$.
\\
From Banknote dataset:\\
partial knowledge rule: $variance > -3.33 and skewness > 5.80$; class 0\\
user corrected rule: $variance > -3.31 and skewness > 5.82$; class 0\\
modifications: changed variance value from $-3.33$ to $-3.31$; changed skewness value from $5.80$ to $5.82$.

\subsection{Rule examples}
Example of Partial Knowledge rule set for the Tic-Tac-Toe domain:
\begin{itemize}
  \item negative class (X does not win):
  \begin{itemize}
    \item $tl \eq ``b'' \land mr \neq ``b'' \land mm \eq ``o'' \land ml \eq ``x'' \land bm \neq ``x'$
    \item $tl \eq ``b'' \land mr \eq ``o'' \land ml \eq ``x'' \land br \eq ``o''$
    \item $tl \eq ``o'' \land mm \eq ``o'' \land ml \eq ``x'' \land br \eq ``o''$
    \item $tl \eq ``o'' \land mr \eq ``x'' \land ml \neq ``b'' \land br \neq ``b'' \land bm \neq ``o'' \land bl \eq ``o''$
    \item $tl \eq ``o'' \land mr \eq ``x'' \land mm \neq ``x'' \land br \eq ``o''$
    \item $tm \eq ``x'' \land tl \neq ``x'' \land mm \eq ``o'' \land bm \neq ``o'' \land bl \eq ``o''$
    \item $tm \eq ``x'' \land tl \eq ``o'' \land ml \neq ``b'' \land br \neq ``b'' \land bl \eq ``o''$
    \item $tr \neq ``x'' \land mm \neq ``x'' \land bl \eq ``o''$
    \item $tr \eq ``o'' \land tm \neq ``b'' \land tl \eq ``o'' \land ml \eq ``x'' \land br \neq ``b'' \land bm \neq ``o''$
    \item $tr \eq ``o'' \land tm \eq ``o'' \land tl \eq ``o'' \land mm \neq ``x''$
  \end{itemize}
  \item positive class (X wins):
  \begin{itemize}
    \item $br \neq ``o'' \land bm \eq ``x'' \land bl \eq ``x''$
    \item $mm \eq ``x''$
    \item $mr \neq ``x'' \land ml \neq ``x'' \land br \neq ``x''$
    \item $tl \eq ``x'' \land bl \neq ``o''$
    \item $tr \eq ``x'' \land br \neq ``o'' \land bl \neq ``o''$
    \item $tr \eq ``x'' \land tl \eq ``x''$
  \end{itemize}
\end{itemize}

\subsection{Complete Set of Transformation Algorithms}
In order to generate a transformation function which satisfies equations 1 or 2, the differences between the explanation clause $e$ and the user feedback clause $e'$ must be identified first. Algorithm~\ref{alg:DiffFunction} shows the procedure for retrieving the literals that differ between the two clauses. As first step all the literals from the two clauses are extracted and stored in two sets (Lines 1-2). Each literal in $e$ that does not appear in $e'$ is stored in a map and the variable in the literal is used as key (Lines 5-7). Another map stores the literals in $e'$ which do not appear in $e$ (Lines 8-10). 
\begin{algorithm} 
\small
\DontPrintSemicolon
 \KwInput{
 \textit{$e$}: explanation clause, \\ 
 \textit{$e'$}: user feedback clause \\ 
 }
 \KwOutput{
  \textit{$\mathbf{map\_1}$}: map $variable \to literal$ where $variable = literal.variable$, $literal \in e$ and $literal \notin e'$,\\
  \textit{$\mathbf{map\_2}$}:  map $variable \to literal$ where $variable = literal.variable$, $literal \in e'$ and $literal \notin e$\\
 }

    $literals\_1 \gets \mathbf{extract\_literals}(e)$\\
    $literals\_2 \gets \mathbf{extract\_literals}(e')$\\
    $map\_1 \gets \mathbf{newMap}()$ \\
    $map\_2 \gets \mathbf{newMap}()$ \\

    $diff\_1 \gets literals\_1 - literals\_2$ \\
    \For{each $literal$ in $diff\_1$} {
        $map\_1.put(literal.variable, literal)$ \\
    }
    
    $diff\_2 \gets literals\_2 - literals\_1$ \\
    \For{each $literal$ in $diff\_2$} {
        $map\_2.put(literal.variable, literal)$ \\
    }

    return $map\_1, map\_2$\\
\caption{DiffFunction.}
\label{alg:DiffFunction}
\end{algorithm} 

A transformation function is implemented as a sequence of operations (or actions) each one targeting a different variable.
Algorithm~\ref{alg:BuildTransformation} depicts how a transformation function is built. The algorithm starts by computing the difference between the two clauses $e$ and $e'$ (Line 4). A transformation action is built for each pair of literals in $e'$ and $e$ that target the same variable but have a different comparison operator and/or value (Lines 5-11). Another set of actions is built for each literal in $e$ whose variable does not appear in any literal of $e'$ (Lines 12-18). Finally, a new transformation function is created which connects together all the generated actions (line 19).
\begin{algorithm} 
\small
\DontPrintSemicolon
 \KwInput{
 \textit{$R$}: original explanation rule, \\ 
 \textit{$R'$}: user feedback rule, \\ 
 \textit{t\_config}: map containing the domains of categorical features and the margin values for numerical features \\
 }
 \KwOutput{
 \textbf{\textit{t}}: transformation function\\
 }

    $actions \gets \mathbf{newList}()$\\
    $e, p \gets R$ \\
    $e', p' \gets R'$ \\
    $map\_1, map\_2 \gets \mathbf{DiffFunction}(e, e')$ \\
    \For{each $variable, literal\_2$ in $map\_2$} {
       \If{$map\_1.has\_key(variable)$}{
            $literal\_1 \gets map\_1.get(variable)$ \\
            $var\_config  \gets t\_config.get(variable)$ \\
            $label\_changed \gets p \neq p'$\\
            $action \gets \mathbf{BuildAction}$ $(literal\_1, literal\_2, var\_config, label\_changed)$\\
            $actions.append(action)$\\
       }
    }
    
    \For{each $variable, literal\_1$ in $map\_1$} {
       \If{\textbf{not} $map\_2.has\_key(variable)$}{
            $var\_config \gets t\_config.get(variable)$ \\
            $label\_changed \gets p \neq p'$\\
            $literal\_2 \gets \emptyset$ \\ 
            $action \gets \mathbf{BuildAction}$ $(literal\_1, literal\_2, var\_config, label\_changed)$\\
            $actions.append(action)$\\
       }
    }

    return $\mathbf{newTransformation}(actions)$\\
\caption{Build Transformation.}
\label{alg:BuildTransformation}
\end{algorithm} 

Given an explanation rule $R$ and a user feedback rule $R'$ there exists tens of possible combinations for the tuple (explanation comparison operator, feedback comparison operator, explanation value, feedback value, model label, feedback label). Each one of these combinations generates a different action function. As most of these functions share the same structure we implemented them as a higher order function that takes as input the decision logic that determines whether the value of a variable should be changed as well as the new value that should be assigned to the variable. Algorithm~\ref{alg:actionFlow} describes the execution flow of an action function. The method \textbf{BuildAction} that appears in Lines 10 and 16 of Algorithm~\ref{alg:BuildTransformation} uses the two literals $literal\_1$ and $literal\_2$ as well as the variable domain or the margin value in $var\_config$ and the boolean flag $label\_changed$ for generating the boolean expressions $c_1$ and $c_2$ and computing the two values $v_1$ and $v_2$. Furthermore, the input parameter $feature_j$ corresponds to the $variable$ in the two literals. For instance, given an explanation clause $e$ as $age \leq 30 \land education = ``Bachelors" \land marital{\text -}status = ``Never{\text -}married"$ and a user correction clause $e'$ as $age > 26 \land education = ``Bachelors" \land marital{\text -}status = ``Never{\text -}married"$ with $p = p'$ then $feature_j$ is the variable $age$, the pair $\langle c_1, v_1 \rangle$ is defined as $\langle value \to value \leq 26, 30 + margin \rangle$ whereas the pair $\langle c_2, v_2 \rangle$ is defined as $\langle value \to value \geq 30, 26 - margin \rangle$. Similarly, given the domain of variable marital-status as $E = \{$Married-civ-spouse, Divorced, Never-married, Separated, Widowed, Married-spouse-absent, Married-AF-spouse$\}$ and a feedback rule $e'$ as $age \leq 30 \land education = ``Bachelors" \land marital{\text -}status \neq ``Divorced"$ with $p = p'$ then, in that case, $feature_j$ is the variable $marital{\text -}status$, the pair $\langle c_1, v_1 \rangle$ is defined as $\langle value \to value \neq ``Divorced", ``Never{\text -}married" \rangle$ whereas the second pair $\langle c_2, v_2 \rangle$ is defined as $\langle value \to False, \emptyset \rangle$.

\begin{algorithm}
\small
\DontPrintSemicolon
 \KwInput{
 \textit{$x_i'$}: data instance to transform, \\
 \textit{$feature_j$}: j-th feature of $X$, \\
 \textit{$c_1$}: boolean expression 1, \\ 
 \textit{$c_2$}: boolean expression 2, \\ 
 \textit{$v_1$}: new value to assign to $feature_j$ if $x_i'.feature_j$ satisfies expression $c_1$, \\
 \textit{$v_2$}: new value to assign to $feature_j$ if $x_i'.feature_j$ satisfies expression $c_2$ \\
 }
    \If{$c_1(x_i'.feature_j) = True$}{
        $x_i'.feature_j \gets v_1$ \\
    }\ElseIf{$c_2(x_i'.feature_j) = True$}{
        $x_i'.feature_j \gets v_2$ \\
    }
\caption{Action function.}
\label{alg:actionFlow}
\end{algorithm}  

Finally, the execution of a transformation function simply involves applying the sequence of actions to a copy of the original instance $x_i$ as detailed in Algorithm~\ref{alg:applyTransformation}. Notice that $feature_j$, the two boolean expressions $c_1$ and $c_2$ and the two values $v_1$ and $v_2$ are bound to the action at the time the action is created in the \textbf{BuildAction} method, therefore when an action function is executed it receives as input parameter only the data instance $x_i'$ (Line 3 of Algorithm~\ref{alg:applyTransformation}).

\begin{algorithm}
\small
\DontPrintSemicolon
 \KwInput{
 \textit{$x_i$}: data input instance, \\ 
 \textit{$t$}: transformation function \\ 
 }
 \KwOutput{$\mathbf{x_i'}$: transformed instance}
    
    $x_i' \gets \mathbf{copy}(x_i)$\\

    \For{each $action$ in $t.actions$}{
       $action.apply(x_i')$ \\
    }
    
    return $x_i'$\\
\caption{Apply Transformation.}
\label{alg:applyTransformation}
\end{algorithm}

\subsection{Experiment Algorithm}
In this section we provide a detailed description of our experiments.

\begin{algorithm}
\small
\DontPrintSemicolon
 \KwInput{
 \textit{FKRS}: rule sets generated using the whole data, \\ 
 \textit{X}: dataset, \\
 \textit{y}: ground truth labels, \\
 \textit{t\_config}: configuration used by the transformation function such as domains of categorical features and margin values for numerical features, \\
 \textit{seed}: value used for generating a shuffle of the data, \\
 \textit{experiment}: flag indicating whether it is running experiment 1 or 2\\
 }
 \KwOutput{$\mathbf{acc\_values}$: accuracy values measured on the hold-out set}
    $X\_train, X\_test, y\_train, y\_test \gets \mathbf{train\_test\_split}(X, y, 0.8, seed)$ \\
    $partitions \gets \mathbf{partition}(X\_train, y\_train, 4)$ \\
    $overlay \gets null$ \\
    $acc\_values \gets \{\}$ \\
    \For{$k \gets 1$ to $3$} {
        $X\_slice, y\_slice \gets \mathbf{concatenate}$(first k partitions) \\
        $pk\_model \gets \mathbf{new\_ml\_model}()$\\
        $pk\_model.train(X\_slice, y\_slice)$\\
        \If{experiment = 1 OR (experiment = 2 AND overlay = null)}{
            $overlay \gets \mathbf{new\_overlay}(pk\_model)$ \\
            $overlay.learn\_model\_rules(X\_slice)$ \\
        }
        $acc \gets \mathbf{compute\_accuracy}$ $(overlay, pk\_model, X\_test, y\_test)$ \\
        $num\_instances \gets X\_slice.size$ \\
        $acc\_values.append((num\_instances, acc))$ \\
        $learning\_slice \gets partitions[k + 1]$\\
        \For{$i \gets 1$ to $learning\_slice.size$} {
            $instance, label \gets learning\_slice[i]$\\
            $fk\_clauses \gets FKRS.explain(instance, label)$ \\
            \If{fk\_clauses is not empty}{
                $response \gets overlay$ $.generate\_response(instance)$ \\
                \If{response does not contain $p'$}{
                    $p, e \gets response$ \\
                    $p' \gets label$ \\
                    $e' \gets \mathbf{random\_select}(fk\_clauses)$ \\
                    \If{e = null}{
                        $overlay.add\_feedback\_rule(e', p', p)$ \\
                    }
                    \Else{
                        \If{$e \neq e'$ OR $p \neq p'$}{
                            $overlay.add\_feedback\_rule$ $(e', e, p', p, t\_config)$ \\
                        }
                    }
                }
            }
            \If{(i mod ($learning\_slice.size/10$)) = 0}{
                $acc \gets \mathbf{compute\_accuracy}$ $(overlay, pk\_model, X\_test, y\_test)$\\
                $num\_instances \gets i + X\_slice.size$ \\
                $acc\_values.append((num\_instances, acc))$ \\
            }
        }
    }
    return $acc\_values$
\caption{Simulation for Experiment 1 and 2}
\label{alg:ExperimentOneAndTwo}
\end{algorithm}  

The simulation for experiment 1 and 2 is described in Algorithm~\ref{alg:ExperimentOneAndTwo}. It starts by splitting the dataset into the training and testing sets with a different split generated each time by means of the seed (Line 1). The training set is further split into 4 partitions (Line 2). The first $k$ partitions are concatenated together at Line 6 and the resulting set is used for training a new ML model (Line 8). Experiment 2 differs from experiment 1 as the overlay solution uses a model trained only on the first partition (Lines 9-11). The algorithm proceeds by iterating over all the instances of the next partition (Lines 15-17). For each instance one or more explanation clauses are retrieved from the FKRS (Line 18). The $generate\_response$ method of the overlay solution (described in the \textit{GenerateResponse} algorithm) is called on each instance as well (Line 20). If no feedback rule applied to the current instance (Line 21) then the clauses from the FKRS are used to simulate the user correcting the framework prediction and/or explanation. If the overlay solution does not provide an explanation clause (Line 25) then a new entry is added to the look-up table with an empty explanation clause $e$. Entries that do not have an explanation clause $e$ are evaluated as a complementary feedback rules (Lines 8-9 of GenerateResponse algorithm). On the other hand, if the overlay solution provides an explanation clause, and it differs from the one provided by FKRS, then a new entry is added to the look-up table (Lines 28-29). Every 10\% of processed data the accuracy of the overlay solution and ML model is computed against the hold-out set (Lines 30-33). Finally, the complete list of accuracy scores is returned (Line 34).

The overall simulation is repeated 50 times using a different seed each time.

\begin{algorithm}
\small
\DontPrintSemicolon
 \KwInput{
 \textit{FKRS}: rule sets generated using the whole data, \\ 
 \textit{X}: dataset, \\
 \textit{y}: ground truth labels, \\
 \textit{t\_config}: map containing the domains of categorical features and the margin values for numerical features (used by the transformation function), \\
 \textit{seed}: value used for generating a shuffle of the data, \\
 \textit{batch\_size}: number of instances to select at each interaction, \\
 \textit{n\_iterations}: number of interactions\\
 }
 \KwOutput{$\mathbf{acc\_values}$: accuracy values measured on the hold-out set}
    $train\_size \gets X.size/50$ \\
    $X\_train, X\_test, y\_train, y\_test \gets \mathbf{train\_test\_split}(X, y, 0.8, seed)$ \\
    $X\_labelled \gets$ a slice of $train\_size$ from $X\_train$ \\
    $y\_labelled \gets$ a slice of $train\_size$ from $y\_train$ \\
    $X\_pool \gets X\_train - X\_labelled$ \\
    $y\_pool \gets y\_train - y\_labelled$ \\
    $selector \gets$ $\mathbf{new\_low\_margin\_selector}()$\\
    $pk\_model \gets \mathbf{new\_ml\_model}()$\\
    $pk\_model.train(X\_labelled, y\_labelled)$\\
    $overlay \gets \mathbf{new\_overlay}(pk\_model)$ \\
    $overlay.learn\_model\_rules(X\_labelled)$ \\
    $acc \gets \mathbf{compute\_accuracy}$ $(overlay, pk\_model, X\_test, y\_test)$ \\
    $acc\_values \gets \{(0, acc)\}$\\
    \For{$k \gets 1$ to $n\_iterations$} {
        $X\_pool\_proba \gets pk\_model.predict\_proba(X\_pool)$ \\
        $X\_batch, y\_batch \gets selector.$ $query(batch\_size, X\_pool, X\_pool\_proba, y\_pool)$ \\
        \For{$i \gets 1$ to $batch\_size$} {
            $instance, label \gets X\_batch[i], y\_batch[i]$\\
            $fk\_clauses \gets FKRS.explain(instance, label)$ \\
            \If{fk\_clauses is not empty}{
                $response \gets overlay$ $.generate\_response(instance)$ \\
                \If{response does not contain $p'$}{
                    $p, e \gets response$ \\
                    $p' \gets label$ \\
                    $e' \gets \mathbf{random\_select}(fk\_clauses)$ \\
                    \If{e = null}{
                        $overlay.add\_feedback\_rule(e', p', p)$ \\
                    }
                    \Else{
                        \If{$e \neq e'$ OR $p \neq p'$}{
                            $overlay.add\_feedback\_rule$ $(e', e, p', p, t\_config)$ \\
                        }
                    }
                }
            }
        }
        $X\_labelled, y\_labelled \gets \mathbf{concatenate}$ $(X\_labelled, y\_labelled, X\_batch, y\_batch)$ \\
        $X\_pool.remove(X\_batch)$\\
        $y\_pool.remove(y\_batch)$\\
        $pk\_model \gets \mathbf{new\_ml\_model}()$\\
        $pk\_model.train(X\_labelled, y\_labelled)$\\
        $acc \gets \mathbf{compute\_accuracy}$ $(overlay, pk\_model, X\_test, y\_test)$ \\
        $acc\_values.append((batch\_size \times k, acc))$\\
    }
    return $acc\_values$ 
\caption{Active Learning Comparison}
\label{alg:ExperimentAL}
\end{algorithm}  

Algorithm~\ref{alg:ExperimentAL} details the simulation for the comparison of our overlay solution with an active learning approach. The data is shuffle and then split into the training and testing sets (Line 2). A slice of size 2\% of the whole data (Line 1) is taken from the training set (Lines 3-4) and used for training the ML model (Line 9). The remaining part of the training set is used as a pool from where extracting the instances to be labelled by the oracle (Line 5-6). For $n\_iterations$ times the following steps are repeated: 1) the \textit{selector} draws a batch of $batch\_size$ instances from the training set (Line 16). 2) the selected instances are used by the overlay solution for learning new rules (Lines 18-30) and eventually retraining the ML model (Line 35). 3) after a batch is processed a new accuracy score is computed against the test set (Line 37). Finally, the complete list of accuracy scores is returned (Line 38).

The overall simulation is repeated 50 times using a different seed each time.

\subsection{Comparison with alternative ML algorithms}

In this section we include additional results for Experiment 1 with the following additional algorithms: DecisionTree, SGDClassifier and RandomForestClassifier.

\begin{figure*}[!t]
\captionsetup[subfigure]{justification=centering}
\centering
\includegraphics[width=1\textwidth]{AuthorKit20/LaTeX/images/finalcharts/decision-tree-all-datasets.png}
 \vspace*{-3mm}
\caption{Experiment 1: Demonstrating Interactive Overlay Approach With a Decision Tree as Machine Learning Model underneath. Lines represent the medians whereas the shadows are the areas between the 25 and 75 percentiles.}
\label{fig:dt}
\end{figure*}

\begin{figure*}[!t]
\captionsetup[subfigure]{justification=centering}
\centering
\includegraphics[width=1\textwidth]{AuthorKit20/LaTeX/images/finalcharts/stochastic-gradient-descent-all-datasets.png}
 \vspace*{-3mm}
\caption{Experiment 1: Demonstrating Interactive Overlay Approach With a Stochastic Gradient Descent Classifier underneath. Lines represent the medians whereas the shadows are the areas between the 25 and 75 percentiles.}
\label{fig:sgd}
\end{figure*}

\begin{figure*}[!t]
\captionsetup[subfigure]{justification=centering}
\centering
\includegraphics[width=1\textwidth]{AuthorKit20/LaTeX/images/finalcharts/random-forest-all-datasets.png}
 \vspace*{-3mm}
\caption{Experiment 1: Demonstrating Interactive Overlay Approach With a Random Forest Classifier underneath. Lines represent the medians whereas the shadows are the areas between the 25 and 75 percentiles.}
\label{fig:rf}
\end{figure*}
As can be seen in the charts of Figures~\ref{fig:dt}, \ref{fig:sgd} and \ref{fig:rf} the trend holds for these three Ml algorithms as was seen for the LogisticRegression model in the paper.

\subsection{Machine Learning Models configuration parameters}

In this section we report the configuration parameters of the different classifiers used in our experiments. Whether a parameter value is not specified, the default value defined in the library implementing the algorithm is used.
\begin{itemize}
    \item BRCG Light\footnote{AI Explainability 360 (v0.2.0)}: lambda0=0.0001, lambda1=0.001
    \item LogisticRegression\footnote{\label{scikit}scikit-learn (v0.23.1)}: max\_iter=500
    \item SGDClassifier\footnotemark[\ref{scikit}]: loss=``log'', max\_iter=500
    \item DecisionTreeClassifier\footnotemark[\ref{scikit}]: criterion=``entropy'', max\_depth=5
    \item RandomForestClassifier\footnotemark[\ref{scikit}]: only default values.
\end{itemize}